%% file: fmow_main.tex
\documentclass[10pt,twocolumn,letterpaper]{article}
\pdfoutput=1

\usepackage{cvpr}
\usepackage{times}
\usepackage{epsfig}
\usepackage{graphicx}
\usepackage{amsmath}
\usepackage{amssymb}
\usepackage{booktabs}
\usepackage{paralist}
\usepackage{textcase}
\usepackage{float}
\usepackage{stackengine}
\usepackage{subcaption}
\usepackage[pagebackref=true,breaklinks=false,colorlinks=true,bookmarks=false]{hyperref}
\usepackage[super]{nth}

\cvprfinalcopy

\begin{document}

\input{definitions}

\title{Functional Map of the World}

\author{Gordon Christie$^1$ \hspace{1.2cm} Neil Fendley$^1$ \hspace{1.2cm} James Wilson$^2$ \hspace{1.2cm} Ryan Mukherjee$^1$ \\ 
$^1$The Johns Hopkins University Applied Physics Laboratory \quad $^2$DigitalGlobe \\  
{\tt \{\href{mailto:gordon.christie@jhuapl.edu}{gordon.christie},\href{mailto:neil.fendley@jhuapl.edu}{neil.fendley},\href{mailto:ryan.mukherjee@jhuapl.edu}{ryan.mukherjee}\}@jhuapl.edu} \\ \tt \href{mailto:james.wilson@digitalglobe.com}{james.wilson@digitalglobe.com} }

\maketitle

\begin{abstract}
We present a new dataset, Functional Map of the World (\fmow), which aims to inspire the development of machine learning models capable of predicting the functional purpose of buildings and land use from temporal sequences of satellite images and a rich set of metadata features. The metadata provided with each image enables reasoning about location, time, sun angles, physical sizes, and other features when making predictions about objects in the image. Our dataset consists of over 1 million images from over 200 countries\footnote{\fmow contains 1,047,691 images covering 207 of the total 247 ISO Alpha-3 country codes.}. For each image, we provide at least one bounding box annotation containing one of 63 categories, including a ``false detection'' category. We present an analysis of the dataset along with baseline approaches that reason about metadata and temporal views. Our data, code, and pretrained models have been made publicly available.
\end{abstract}

\input{introduction}
\input{related_work}

\input{dataset_collection}
\input{dataset_analysis}
\input{baselines_methods}
\input{conclusion_discussion}

\input{appendix.tex}

\bibliographystyle{ieee}
\bibliography{references}

\end{document}

%% file: definitions.tex
\newcommand{\figref}[1]{Figure~\ref{#1}}
\newcommand{\secref}[1]{Section~\ref{#1}}
\newcommand{\fmow}{fMoW\xspace}

\newcommand*\rot{\rotatebox{90}}

\newcommand{\cnnavg}{\textsc{CNN-IM}\xspace}
\newcommand{\cnnavgI}{\textsc{CNN-I}\xspace}
\newcommand{\lstm}{\textsc{LSTM-IM}\xspace}
\newcommand{\lstmI}{\textsc{LSTM-I}\xspace}
\newcommand{\lstmM}{\textsc{LSTM-M}\xspace}

\newcommand{\train}{\texttt{train}\xspace}
\newcommand{\val}{\texttt{val}\xspace}
\newcommand{\test}{\texttt{test}\xspace}
\newcommand{\seq}{\texttt{seq}\xspace}

\newcommand{\subsubheaderbf}[1]{\mbox{\textbf{#1}\hspace*{2.5mm}}}
\newcommand{\subsubheaderit}[1]{\mbox{\textit{#1}\hspace*{2.5mm}}}
\newcommand{\subsubheadertt}[1]{\mbox{\texttt{#1}\hspace*{2.5mm}}}
\newcommand{\subsubheader}[1]{\mbox{#1\hspace*{2.5mm}}}

\newcommand{\fmowfull}{\texttt{\fmow-full}\xspace}
\newcommand{\fmowrgb}{\texttt{\fmow-rgb}\xspace}

%% file: introduction.tex
\vspace{-0.5cm}

\section{Introduction}
\label{sec:introduction}

Satellite imagery presents interesting opportunities for the development of object classification methods. Most computer vision (CV) datasets for this task focus on images or videos that capture brief moments \cite{russakovsky2015imagenet,lin2014microsoft}. With satellite imagery, temporal views of objects are available over long periods of time. In addition, metadata is available to enable reasoning beyond visual information. For example, by combining temporal image sequences with timestamps, models may learn to differentiate office buildings from multi-unit residential buildings by observing whether or not their parking lots are full during business hours. Models may also be able to combine certain metadata parameters with observations of shadows to estimate object heights. In addition to these possibilities, robust models must be able to generalize to unseen areas around the world that may include different building materials and unique architectural styles.

\begin{figure}[t!]
    \centering
    \includegraphics[width=\columnwidth]{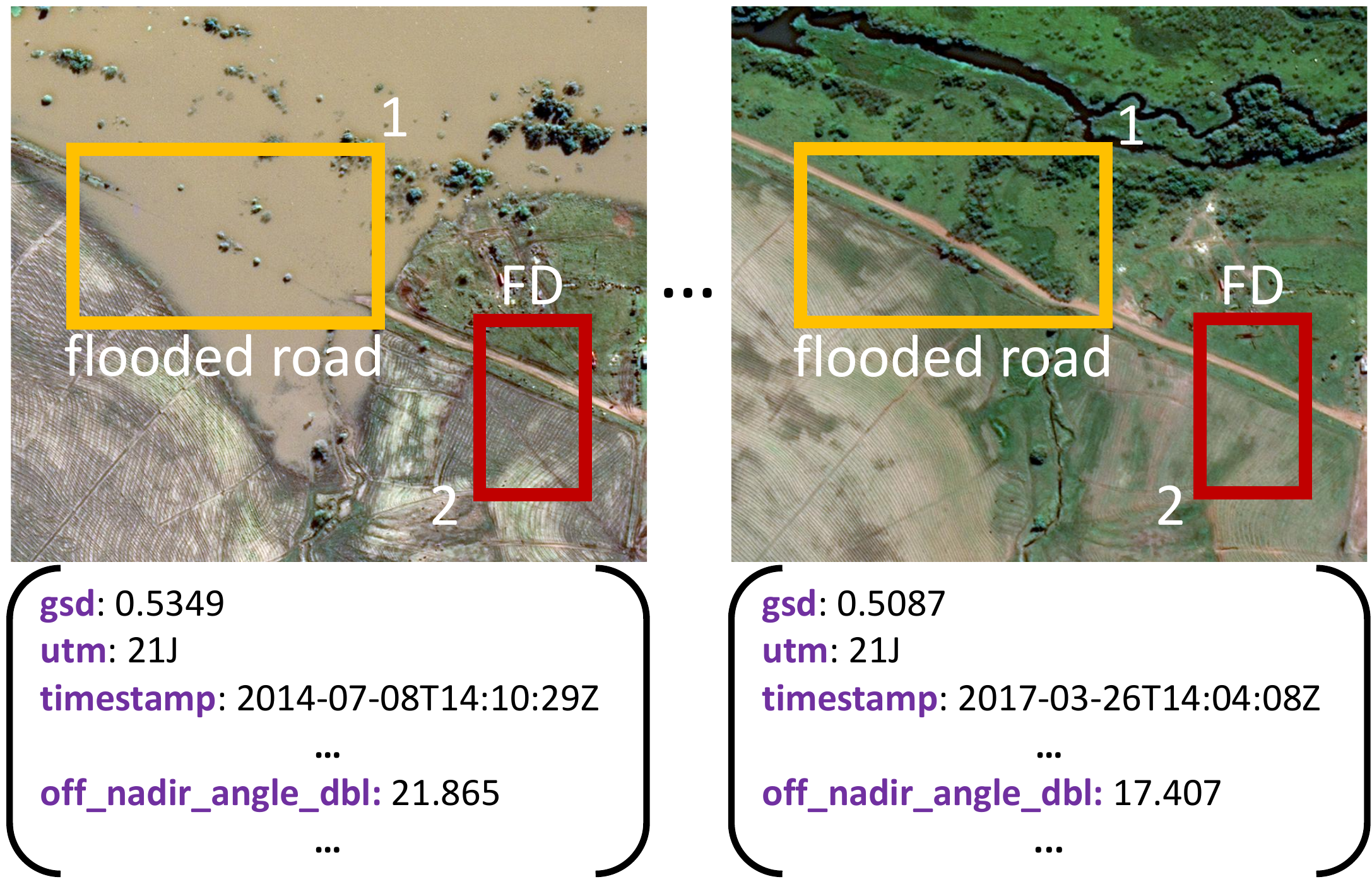}
    \vspace{-0.5cm}
    \caption{In \fmow, temporal sequences of images, multispectral imagery, metadata, and bounding boxes are provided. In this example, if we only look inside the yellow box in the right image, we will only see a road and vegetation. On the other hand, if we only see the water in the left image, then we will potentially predict this to be a lake. However, by observing both views of this area, we can now reason that this sequence contains a flooded road.}
    \label{fig:teaser}
    \vspace{-0.4cm}
\end{figure}

Enabling the aforementioned types of reasoning requires a large dataset of annotated and geographically diverse satellite images. In this work, we present our efforts to collect such a dataset, entitled Functional Map of the World (\fmow). \fmow has several notable features, including global diversity, a variable number of temporal images per scene, multispectral imagery, and metadata associated with each image. The task posed for our dataset falls in between object detection and classification. That is, for each temporal sequence of images, at least one bounding box is provided that maps to one of 63 categories, including a ``false detection'' (FD) category that represents content not characterized by the other 62 categories. These boxes are intended to be used as input to a classification algorithm. \figref{fig:teaser} shows an example.

Collecting a dataset such as \fmow presents some interesting challenges. For example, one consideration would be to directly use crowdsourced annotations provided by OpenStreetMap\footnote{\scriptsize \url{https://www.openstreetmap.org}} (OSM). However, issues doing so include inconsistent, incorrect, and missing annotations for a large percentage of buildings and land use across the world. Moreover, OSM may only provide a single label for the current contents of an area, making it difficult to correctly annotate temporal views. Another possibility is to use the crowd to create annotations from scratch. However, annotating instances of a category with no prior information is extremely difficult in a large globally-diverse satellite image dataset. This is due, in part, to the unique perspective that satellite imagery offers when compared with ground-based datasets, such as ImageNet~\cite{russakovsky2015imagenet}. Humans are seldom exposed to aerial viewpoints in their daily lives. As such, objects found in satellite images tend to be visually unfamiliar and difficult to identify. Buildings can also be repurposed throughout their lifetime, making visual identification even more difficult. For these reasons, we use a multi-phase process that combines map data and crowdsourcing.

Another problem for \fmow is that annotating every instance of a category is made very difficult by the increased object density of certain categories. For example, single-unit residential buildings often occur in dense clusters alongside other categories, where accurately discriminating and labeling every building would be very time-consuming. To address this shortcoming, we propose providing bounding boxes as algorithm input, unlike a typical detection dataset and challenge where bounding boxes are expected as output. This circumvents full image annotation issues that stem from incomplete map data and visual unfamiliarity. As a result, data collection could focus on global diversity and annotations could be limited to a small number of high-confidence category instances per image.

Our contributions are summarized as follows: (1) We provide the largest publicly available satellite dataset containing bounding box annotations, multispectral imagery, metadata, and revisits. This enables joint reasoning about images and metadata, as well as long-term temporal reasoning for areas of interest. (2) We present methods based on CNNs that exploit the novel aspects of our dataset, with performance evaluation and comparisons, which can be applied to similar problems in other application domains. Our code, data, and pretrained models have all been publicly released\footnote{\scriptsize \url{https://github.com/fMoW}}.
In the following sections, we provide an analysis of \fmow and baseline methods for the task.

As an aside, in addition to collecting and publishing \fmow, a public prize challenge\footnote{\scriptsize \url{https://www.iarpa.gov/challenges/fmow.html}} was organized around the dataset. It ran from Sep. 14 - Dec. 31 2017. The top 3 participants have open-sourced their solutions on the \fmow GitHub page. These methods, as well as the baseline, were developed using the publicly available data. However, all data, including the sequestered data used for final testing, is now publicly available. 

%% file: related_work.tex
\section{Related Work}
\label{sec:related_work}

While large datasets are nothing new to the vision community, they have typically focused on first-person or ground-level imagery \cite{russakovsky2015imagenet,lin2014microsoft,abu2016youtube,fei2006one,griffin2007caltech,everingham2015pascal,openimages}. This is likely due in part to the ease with which this imagery can be collected and annotated. Recently, there have been several, mostly successful, attempts to leverage techniques that were founded on first-person imagery and apply them to remote sensing data \cite{jean2016combining,marmanis2016deep,xia2017exploiting}. However, these efforts highlight the research gap that has developed due to the lack of a large dataset to appropriately characterize the problems found in remote sensing.
We now highlight several of these areas where we believe \fmow can make an impact.

\noindent \subsubheaderbf{Reasoning Beyond Visual Information}
Many works have extended CV research to simultaneously reason about other modules of perception~\cite{antol2015vqa,karpathy2015deep,Pan_2016_CVPR,harwath2017learning,chang2017matterport3d}.
In this work, we are interested in supporting joint reasoning about temporal sequences of images and associated metadata features. One of these features is UTM zone, which provides location context. In a similar manner, \cite{tang2015improving} shows improved image classification results by jointly reasoning about GPS coordinates and images, where several features are extracted from the coordinates, including high-level statistics about the population. 
Although we use coarser location features (UTM zones) than GPS in this work, we do note that using similar features would be an interesting study. GPS data for \fmow imagery was also made publicly available after the end of the prize challenge.

\noindent \subsubheaderbf{Multi-view Classification}
Satellite imagery offers a unique and somewhat alien perspective on the world. Most structures are designed for recognition from ground level. 
As such, it can be difficult, if not impossible, to identify functional purpose from a single overhead image.
One of the ways \fmow attempts to address this issue is by providing multiple temporal views of each object, when available. Along these lines, several works in the area of video processing have been able to build upon advancements in single image classification \cite{karpathy2014large,donahue2015long,yue2015beyond} to create networks capable of extracting spatio-temporal features. These works may be a good starting point, but it is important to keep in mind the vastly different temporal resolution on which these datasets operate. 
For example, the YouTube-8M dataset \cite{abu2016youtube} contains videos with 30 frames per second. For satellites, it is not uncommon for multiple days to pass before they can image the same location, and possibly months before they can get an unobstructed view.

Perhaps the most similar work to ours in terms of temporal classification is PlaNet~\cite{weyand2016planet}. They pose the image localization task as a classification problem, where photo albums are classified as belonging to a particular bucket that bounds an area on the globe. We use a similar approach in one of our baseline methods. 

\noindent \subsubheaderbf{Remote Sensing Datasets}
One of the earliest annotated satellite datasets similar to \fmow is the UC Merced Land Use Dataset, which offers 21 categories and 100 images per category with roughly 30cm resolution and image sizes of 256x256 \cite{yang2010bag}. Another recent dataset similar to \fmow is TorontoCity~\cite{wang2016torontocity}, which includes aerial imagery captured during different seasons in the greater Toronto area. While they present several tasks, the two that are similar to land-use classification are zoning classification and segmentation (\eg, residential, commercial). 
Datasets have also been created for challenges centered around semantic segmentation, such as the IEEE GRSS Data Fusion Contests~\cite{debes2014hyperspectral} and the ISPRS 2D Semantic Labeling Contest~\cite{ISPRS2D}.

SpaceNet \cite{spacenet}, a recent dataset that has received substantial attention, contains both 30cm and 50cm data of 5 cities. While it mainly includes building footprints, point of interest (POI) data was recently released into SpaceNet that includes locations of several categories within Rio de Janeiro. Other efforts have also been made to label data from Google Earth, such as the AID \cite{xia2017aid} (10,000 images, 30 categories) and NWPU-RESISC45 (31,500 images of 45 categories) \cite{cheng2017remote} datasets. In comparison, \fmow offers 1,047,691 images of 63 categories, and includes associated metadata, temporal views, and multispectral data, which are not available from Google Earth.

%% file: dataset_collection.tex
\section{Dataset Collection}
\label{sec:dataset_collection}

Prior to the dataset collection process for \fmow, a set of categories had to be identified. Based on our target of 1 million images, collection resources, plan to collect temporal views, and discussions with researchers in the CV community, we set a goal of including between 50 and 100 categories. We searched sources such as the OSM Map Features\footnote{\scriptsize \url{https://wiki.openstreetmap.org/wiki/Map_Features}} list and NATO Geospatial Feature Concept Dictionary\footnote{\scriptsize \url{https://portal.dgiwg.org/files/?artifact_id=8629}} for categories that highlight some of the challenges discussed in Section \ref{sec:related_work}. For example, ``construction site'' and ``impoverished settlement'' are categories from \fmow that may require temporal reasoning to identify, which presents a unique challenge due to temporal satellite image sequences typically being scattered across large time periods. We also focused on grouping categories according to their functional purpose to encourage the development of approaches that can generalize. For example, by grouping recreational facilities (\eg, tennis court, soccer field), algorithms would hopefully learn features common to these types of facilities and be able to recognize other recreational facilities beyond those included in the dataset (\eg, rugby fields). This also helps avoid issues related to label noise in the map data.

Beyond research-based rationales for picking certain categories, we had some practical ones as well. Before categories could be annotated within images, we needed to find locations where we have high confidence of their existence. This is where maps play a crucial role. ``Flooded road'', ``debris or rubble'', and ``construction site'' were the most difficult categories to collect since open source data does not generally contain temporal information. However, with more careful search procedures, reuse of data from humanitarian response campaigns, and calculated extension of keywords to identify categories even when not directly labeled, we were able to collect temporal stacks of imagery that contained valid examples.

All imagery used in \fmow was collected from the DigitalGlobe constellation\footnote{\scriptsize \url{https://www.digitalglobe.com/resources/satellite-information}}.
Images were gathered in pairs, consisting of 4-band or 8-band multispectral imagery in the visible to near-infrared region, as well as a pan-sharpened RGB image that represents a fusion of the high-resolution panchromatic image and the RGB bands from the lower-resolution multispectral image. 4-band imagery was obtained from either the QuickBird-2 or GeoEye-1 satellite systems, whereas 8-band imagery was obtained from WorldView-2 or WorldView-3.

\begin{figure}[h!]
	\includegraphics[width=\columnwidth]{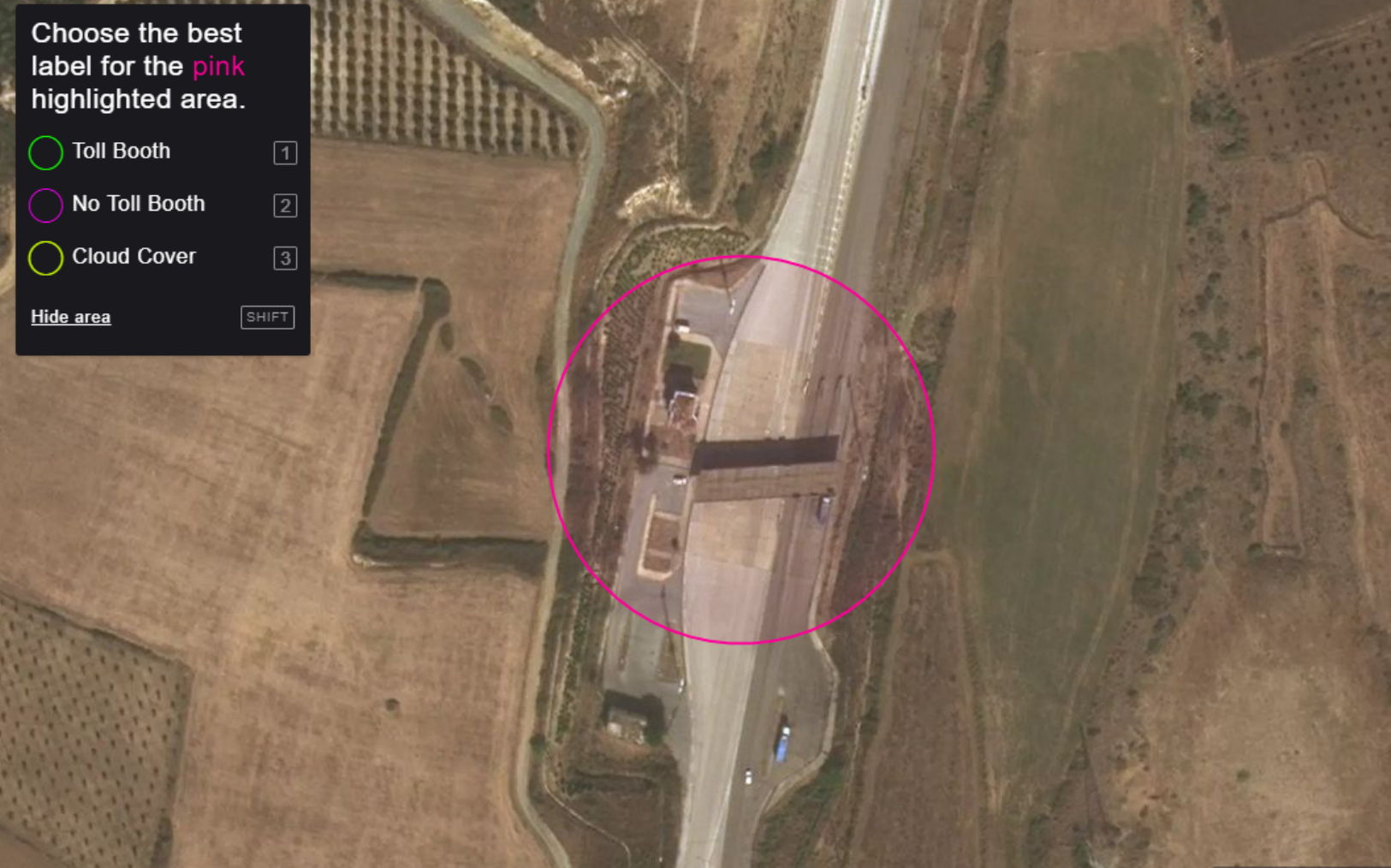}
	\caption{Sample image of what a GeoHIVE user might see while validating potential \fmow dataset features. Instructions can be seen in the top-left corner that inform users to press the `1', `2', or `3' keys to validate existence, non-existence, or cloud obscuration of a particular object.}
	\label{fig:geohive_confirm_cat_example1}
\end{figure}

More broadly, \fmow was created using a three-phase workflow consisting of location selection, image selection, and bounding box creation. The location selection phase was used to identify potential locations that map to our categories while also ensuring geographic diversity. Potential locations were drawn from several Volunteered Geographic Information (VGI) datasets, which were conflated and curated to remove duplicates. To ensure diversity, we removed neighboring locations within a specified distance (typically 500m) and set location frequency caps for categories that have severely skewed geographic distributions. These two factors helped reduce spatial density while also encouraging the selection of locations from disparate geographic areas. The remaining locations were then processed using DigitalGlobe's GeoHIVE\footnote{\scriptsize \url{https://geohive.digitalglobe.com}} crowdsourcing platform. Members of the GeoHIVE crowd were asked to validate the presence of categories in satellite images, as shown in \figref{fig:geohive_confirm_cat_example1}.

The image selection phase comprised of a three-step process, which included searching the DigitalGlobe satellite imagery archive, creating image chips, and filtering out cloudy images. Approximately 30\% of the candidate images were removed for being too cloudy. DigitalGlobe's IPE Data Architecture Highly-available Object-store service was used to process imagery into pan-sharpened RGB and multispectral image chips in a scalable fashion. These chips were then passed through a CNN architecture to classify and remove any undesirable cloud-covered images.

Finally, images that passed through the previous two phases were sent to a curated and trusted crowd for bounding box annotation. This process involved a separate interface from the first phase, where crowd users were asked to draw bounding boxes around the category of interest in each image and were provided some category-specific guidance for doing so. The resulting bounding boxes were then graded by a second trusted crowd to assess quality. The trusted crowd includes individuals from universities and elsewhere that have a strong relationship with DigitalGlobe or the labeling campaigns they have conducted. In total, 642 unique GeoHIVE users required a combined total of approximately 2,800 hours to annotate category instances for \fmow.

Even after multiple crowd validation procedures and implementing programmatic methods for ensuring geographic diversity, there were several categories that contained some bias. For example, the ``wind farm'' category does not contain very many examples from the United States, even though the initial location selection phase returned 1,938 viable locations from the United States. Many of these ``wind farm'' instances were invalidated by the crowd, likely due to the difficulty of identifying tall, thin structures in satellite imagery, particularly when the satellite image is looking straight down on the tower. The ``barn'', ``construction site'', ``flooded road'', and ``debris or rubble'' categories are also examples that contain some geographic bias. In the case of the ``barn'' category, the bias comes from the distribution of ``barn'' tags in OSM, which are predominately located in Europe, whereas the other three categories contain geographic bias as a result of the more complex feature selection process, mentioned earlier, that was required for these categories. FD boxes were included to mitigate this bias. When they are present in an image, algorithms are forced to use the imagery to accurately make predictions, as there may be two boxes with different labels that share similar metadata features.

The following provides a summary of the metadata features included in our dataset, as well as any preprocessing operations that are applied before input into the baseline methods:

\begin{compactitem}
\item \subsubheaderbf{UTM Zone} One of 60 UTM zones and one of 20 latitude bands are combined for this feature. We convert these values to 2 coordinate values, each between 0 and 1. This is done by taking the indices of the values within the list of possible values and then normalizing. While GPS data is now publicly available, it was withheld during the prize challenge to prevent participants from precisely referencing map data.
\item \subsubheaderbf{Timestamp} The year, month, day, hour, minute, second, and day of the week are extracted from the timestamp and added as separate features. The timestamp provided in the metadata files is in Coordinated Universal Time (UTC).
\item \subsubheaderbf{GSD} Ground sample distance, measured in meters, is provided for both the panchromatic and multispectral bands in the image strip. The panchromatic images used to generate the pan-sharpened RGB images have higher resolution than the MSI, and therefore have smaller GSD values. These GSD values, which describe the physical sizes of pixels in the image, are used directly without any preprocessing. 
\item \subsubheaderbf{Angles} These identify the angle at which the sensor is imaging the ground, as well as the angular location of the sun with respect to the ground and image. The following angles are provided:
\begin{compactitem}
\item \subsubheaderit{Off-nadir Angle} Angle in degrees (0-90$^\circ$) between the point on the ground directly below the sensor and the center of the image swath.
\item \subsubheaderit{Target Azimuth} Angle in degrees (0-360$^\circ$) of clockwise rotation off north to the image swath's major axis.
\item \subsubheaderit{Sun Azimuth} Angle in degrees (0-360$^\circ$) of clockwise rotation off north to the sun.
\item \subsubheaderit{Sun Elevation} Angle in degrees (0-90$^\circ$) of elevation, measured from the horizontal, to the sun.
\end{compactitem}
\item \subsubheaderbf{Image+box sizes} The pixel dimensions of the bounding boxes and image size, as well as the fraction of the image width and height that the boxes occupy, are added as features. 
\end{compactitem}

A full list of metadata features and their descriptions can be found in the appendix. 

%% file: dataset_analysis.tex
\section{Dataset Analysis}
\label{sec:dataset_analysis}

\begin{figure*}[t!]
	\centering
	\includegraphics[width=\textwidth]{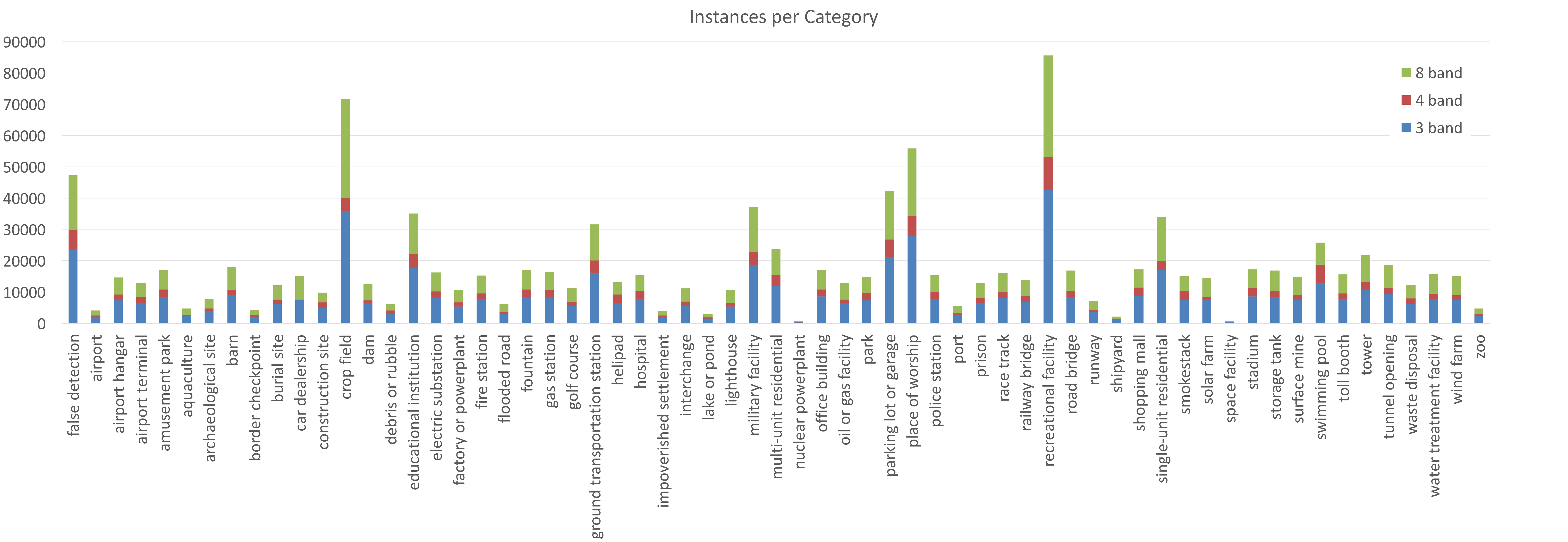}
	\caption{This shows the total number of instances for each category (including FD) in \fmow across different number of bands. These numbers include the temporal views of the same areas. \fmowfull consists of 3 band imagery (pan-sharpened RGB), as well as 4 and 8 band imagery. In \fmowrgb, the RGB channels of the 4 and 8 band imagery are extracted and saved as JPEG images.}
	\label{fig:fmow_category_distribution}
    \vspace{-0.3cm}
\end{figure*}

Here we provide some statistics and analysis of \fmow. Two versions of the dataset are publicly available:

\begin{compactitem}
\item \subsubheadertt{\fmow-full} The full version of the dataset includes pan-sharpened RGB images and 4/8-band multispectral images (MSI), which are both stored in TIFF format. Pan-sharpened images are created by ``sharpening'' lower-resolution MSI using higher-resolution panchromatic imagery \cite{padwick2010worldview}. All pan-sharpened images in \fmowfull have corresponding MSI, where the metadata files for these images are nearly identical.
\item \subsubheadertt{\fmow-rgb} An alternative JPEG compressed version of the dataset, which is provided due to the large size of \fmowfull. For each pan-sharpened RGB image we simply perform a conversion to JPEG. For MSI images, we extract the RGB channels and save them as JPEGs.
\end{compactitem}

For all experiments presented in this paper, we use \fmowrgb. We also exclude RGB-extracted versions of the MSI in \fmowrgb, as they are effectively downsampled versions of the pan-sharpened RGB images.

\subsection{\fmow Splits}

We have made the following splits to the dataset:

\begin{compactitem}
\item \subsubheader{\train} Contains 83,412 (62.85\%) of the total unique bounding boxes.

\item \subsubheader{\val} Contains 14,241 (10.73\%) of the total unique bounding boxes. This set was made representative of \test, so that validation can be performed.

\item \subsubheader{\test} Contains 16,948 (12.77\%) of the total unique bounding boxes.

\item \subsubheader{\seq} Contains 18,115 (13.65\%) of the total unique bounding boxes. This set was also made representative of \test, but was not publicly released during the prize challenge centered around this dataset.
\end{compactitem}

Each split was formed by first binning the GSD, number of temporal views per sequence, UTM zone, and off-nadir angle values. After binning these values, temporal sequences were divided between the different dataset splits while ensuring that the counts for these bins, as well as the distribution of categories per split, were consistent. Singleton sequences, such as those that are the only ones to cover a particular UTM zone, were also evenly distributed between the various splits. The total number of bounding box instances for each category can be seen in \figref{fig:fmow_category_distribution}.

\subsection{\fmow Statistics}

\begin{figure}[h!]
	\centering
	\includegraphics[width=\columnwidth]{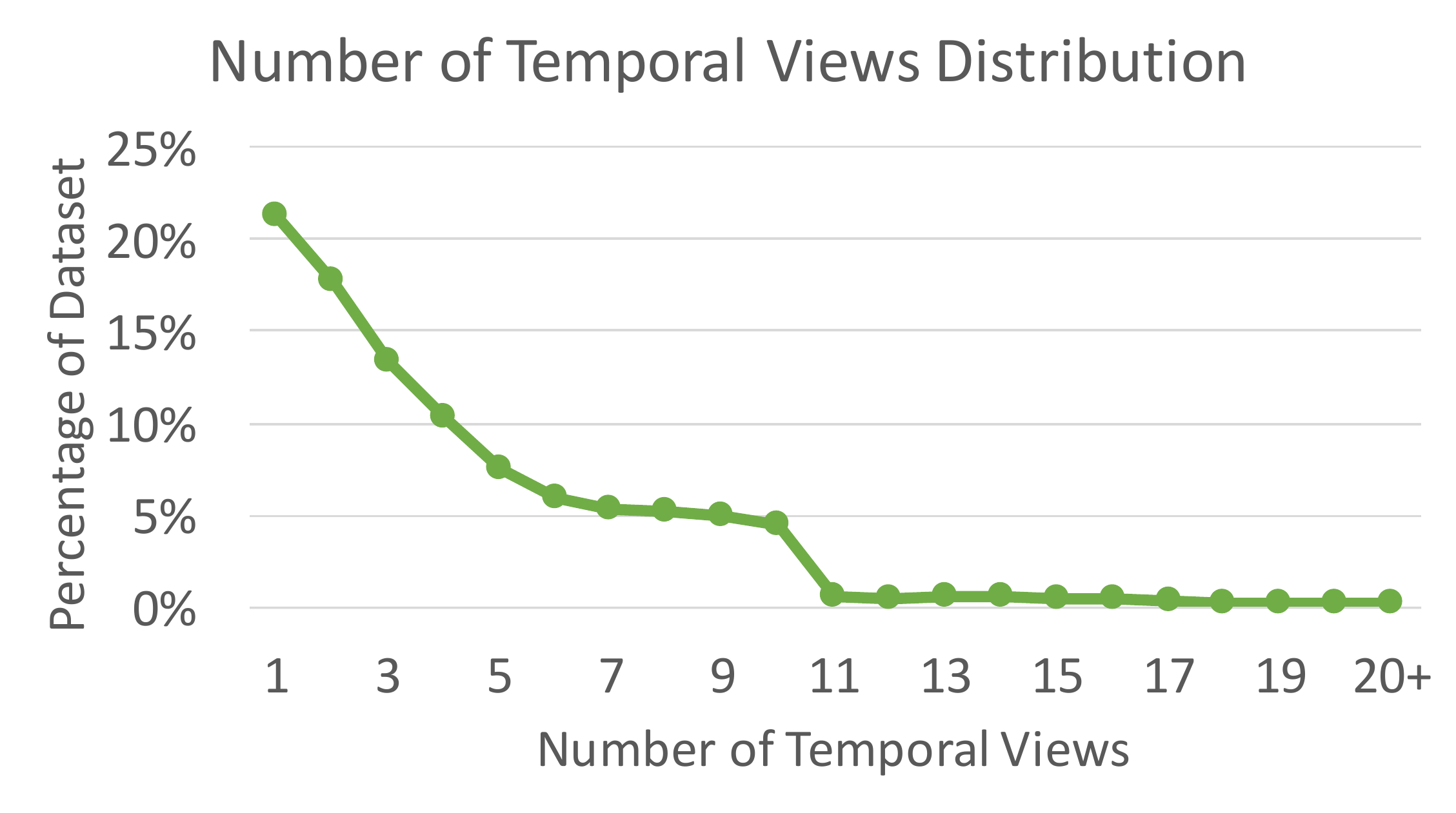}
    \vspace{-0.6cm}
	\caption{This shows the distribution of the number of temporal views in our dataset. The number of temporal views is not incremented by both the pan-sharpened and multispectral images. These images have almost identical metadata files and are therefore not counted twice. The maximum number of temporal views for any area in the dataset is 41.}
	\label{fig:temporal_views_distribution}
\end{figure}

Variable length sequences of images are provided for each scene in the dataset. \figref{fig:temporal_views_distribution} shows the distribution of sequence lengths in \fmow. 21.2\% of the sequences contain only 1 view. Most (95\%) of the sequences contain 10 or fewer images. 

\begin{figure}[h!]
	\centering
	\includegraphics[height=4.5cm]{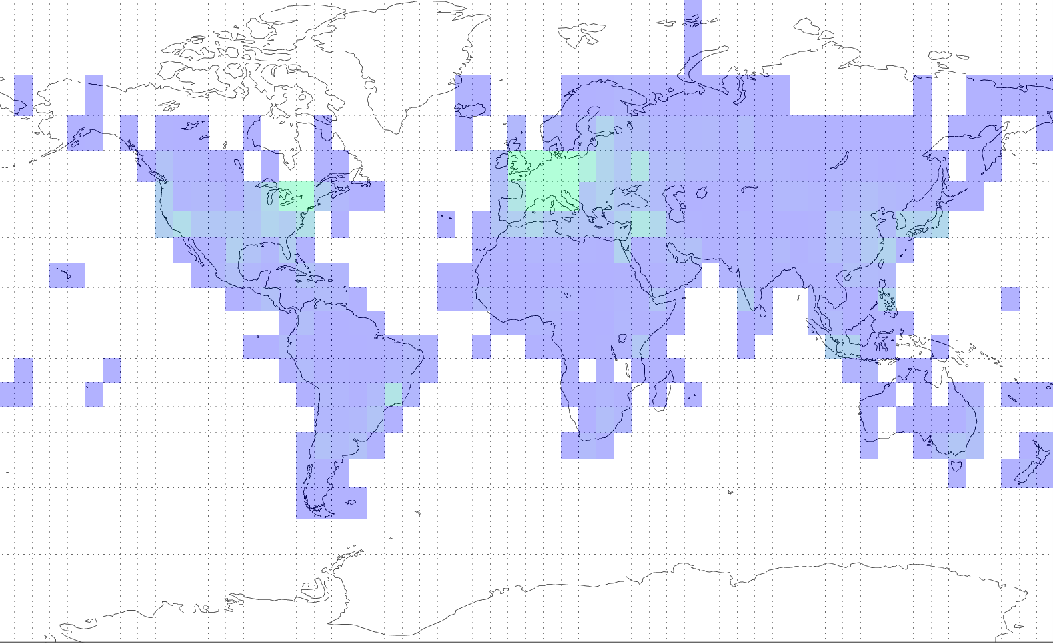}
	\includegraphics[height=4.5cm]{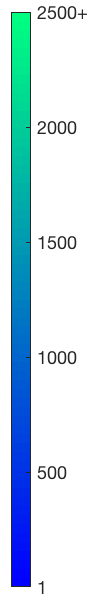}
    \vspace{-0.3cm}
	\caption{This shows the geographic diversity of \fmow. Data was collected from over 400 unique UTM zones (including latitude bands). This helps illustrate the number of images captured in each UTM zone, where more green colors show UTM zones with a higher number of instances, and more blue colors show UTM zones with lower counts.}
	\label{fig:utm_plot}
    \vspace{-0.4cm}
\end{figure}

A major focus of the collection effort was global diversity. In the metadata, we provide UTM zones, which typically refer to 6$^\circ$ longitude bands (1-60). We also concatenate letters that represent latitude bands (total of 20) to the UTM zones in the metadata. \figref{fig:utm_plot} illustrates the frequency of sequences within the UTM zones on earth, where the filled rectangles each represent a different UTM zone. Green colors represent areas with higher numbers of sequences, while blue regions have lower counts. As seen, \fmow covers much of the globe. 

\begin{figure}[h!]
    \centering
    \begin{subfigure}[t]{0.48\columnwidth}
    \centering
    \includegraphics[width=\columnwidth]{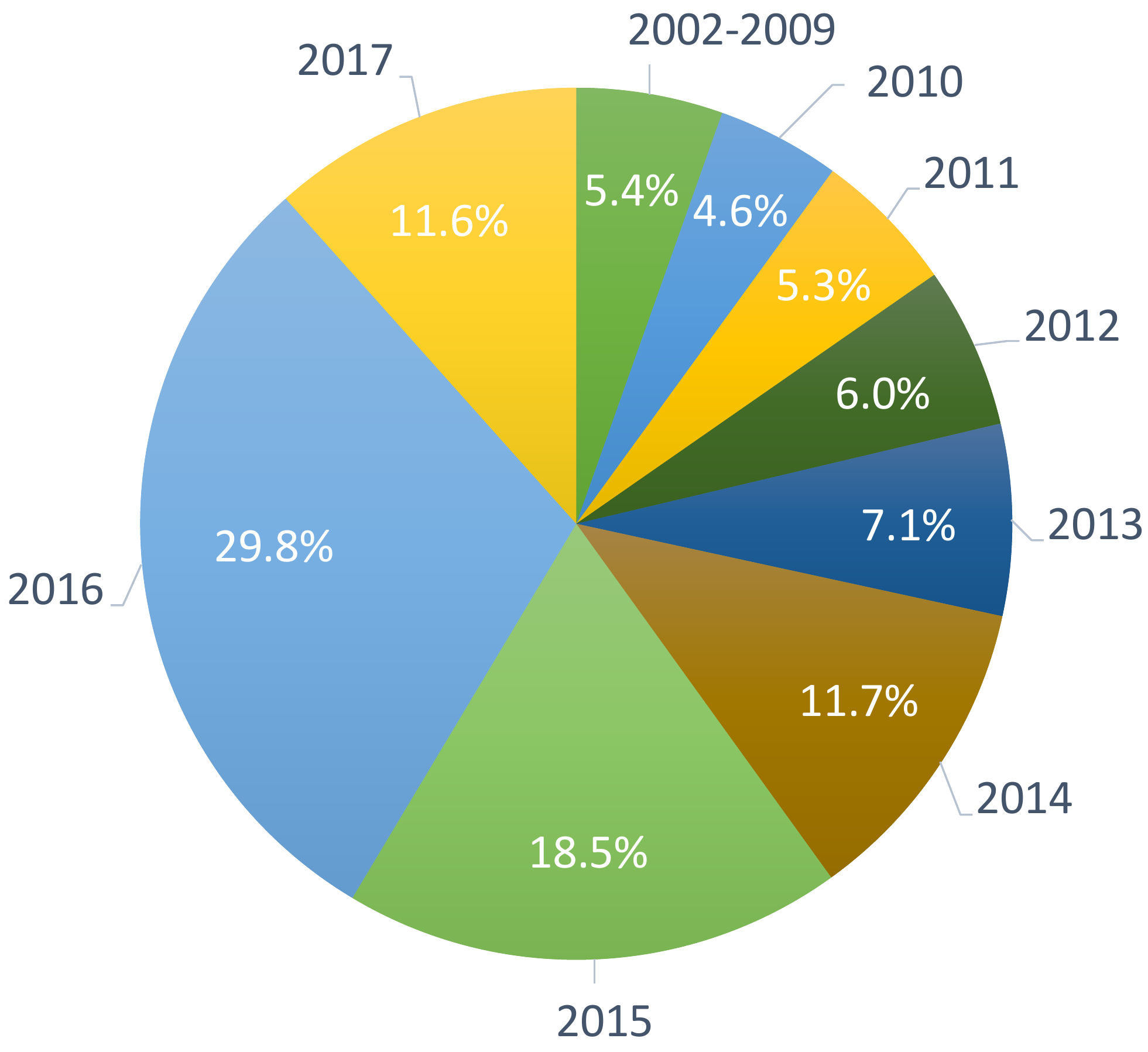}
    \caption{}
    \end{subfigure}
    \begin{subfigure}[t]{0.48\columnwidth}
    \centering
    \includegraphics[width=\columnwidth]{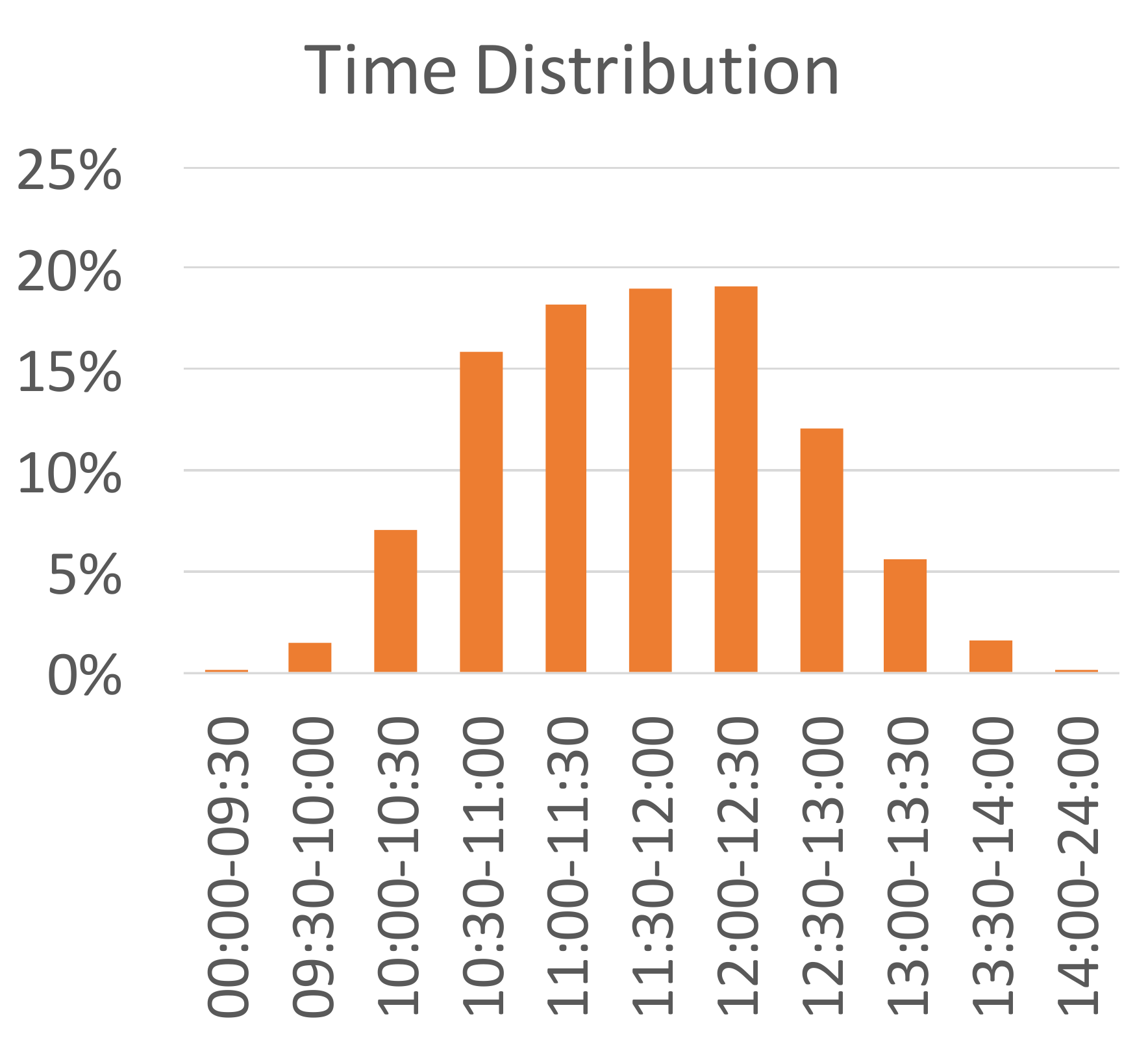}
    \caption{}
    \end{subfigure}
    \vspace{-0.3cm}
    \caption{Distribution over (a) years the images were captured, and (b) time of day the images were captured (UTC converted to local time for this figure).}
    \label{fig:time_distribution}
    \vspace{-0.4cm}
\end{figure}

The images captured for \fmow also have a wide range of dates, which, in some cases, allows algorithms to analyze areas on earth over long periods of time. \figref{fig:time_distribution} shows distributions for years and local times (converted from UTC) in which the images were captured. The average time difference between the earliest and most recent images in each sequence is approximately 3.8 years.

%% file: baselines_methods.tex
\section{Baselines and Methods}
\label{sec:baselines_methods}

Here we present 5 different approaches to our task, which vary by their use of metadata and temporal reasoning. All experiments were performed using \fmowrgb. Two of the methods presented involve fusing metadata into a CNN architecture in an attempt to enable the types of reasoning discussed in the introduction. We perform mean subtraction and normalization for the metadata feature vectors using values calculated over \train + \val. 

It is worth noting here that the imagery in \fmow is not registered, and while many sequences have strong spatial correspondence, individual pixel coordinates in different images do not necessarily represent the same positions on the ground. As such, we are prevented from easily using methods that exploit registered sequences.

The CNN used as the base model in our various baseline methods is DenseNet-161~\cite{huang2017densely}, with 48 feature maps ($k$=48). During initial testing, we found this model to outperform other models such as VGG-16~\cite{simonyan2014very} and ResNet-50~\cite{he2016deep}. We initialize our base CNN models using the pretrained ImageNet weights, which we found to improve performance during initial tests. Training is performed using a crop size of 224x224, the Adam optimizer~\cite{kingma2014adam}, and an initial learning rate of 1e-4. Due to class imbalance in our dataset, we attempted to weight the loss using class frequencies, but did not observe any improvement. 

\begin{figure}[h!]
	\centering
	\includegraphics[width=\columnwidth]{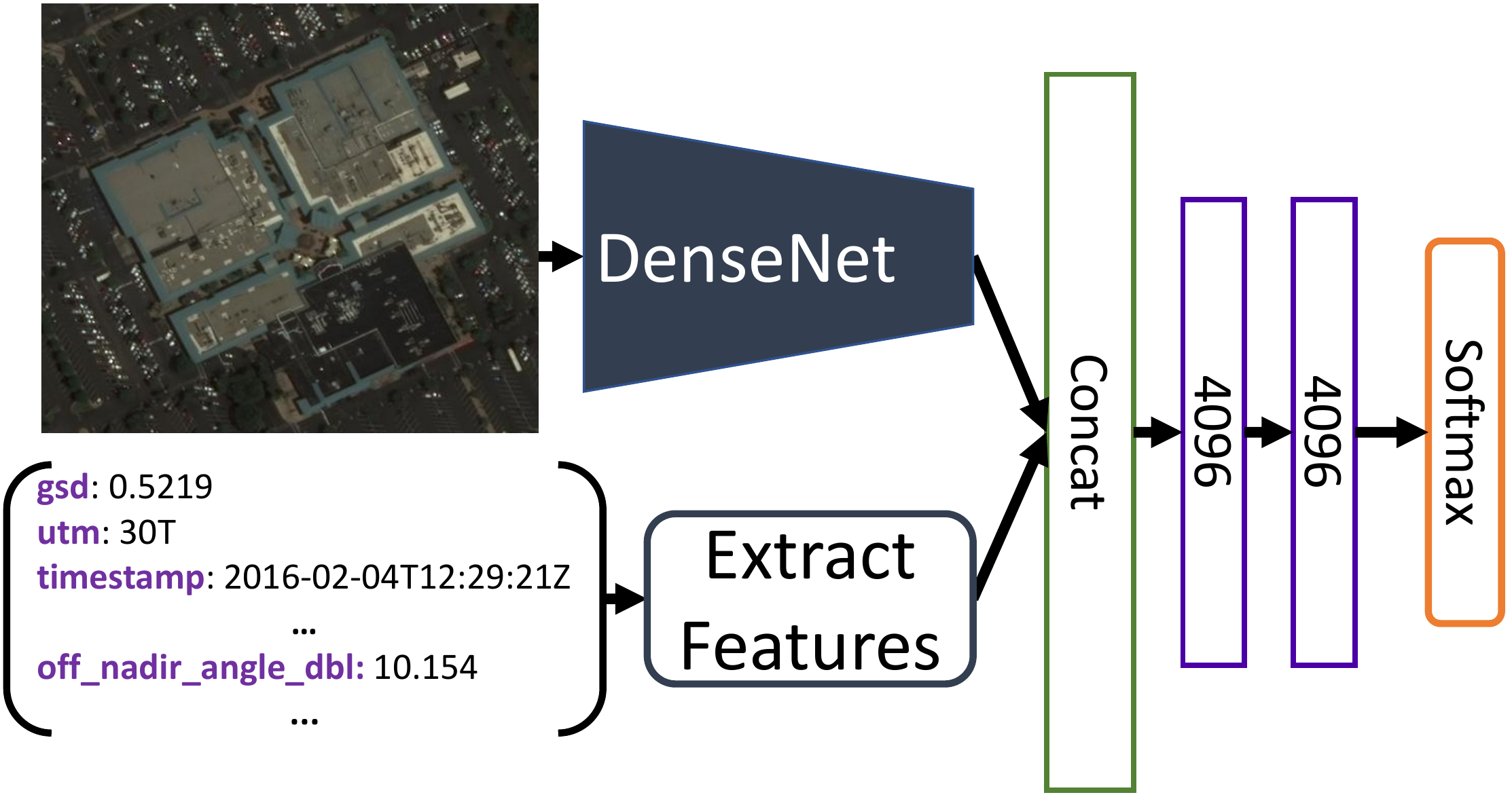}
    \vspace{-0.7cm}
	\caption{An illustration of our base model used to fuse metadata features into the CNN. This model is used as a baseline and also as a feature extractor (without softmax) for providing features to an LSTM. Dropout layers are added after the 4096-d FC layers.}
	\label{fig:baseline_overview}
\end{figure}

To merge metadata features into the model, the softmax layer of DenseNet is removed and replaced with a concatenation layer to merge DenseNet features with metadata features, followed by two 4096-d fully-connected layers with 50\% dropout layers, and a softmax layer with 63 outputs (62 main categories + FD). An illustration of this base model is shown in \figref{fig:baseline_overview}.

We test the following approaches with \fmow:

\begin{compactitem}
\item \subsubheaderbf{\lstmM} An LSTM architecture trained using temporal sequences of metadata features. We believe training solely on metadata helps understand how important images are in making predictions, while also providing some measure of bias present in \fmow.
\item \subsubheaderbf{\cnnavgI} A standard CNN approach using only images, where DenseNet is fine-tuned after ImageNet. Softmax outputs are summed over each temporal view, after which an argmax is used to make the final prediction. The CNN is trained on all images across all temporal sequences of \train + \val.
\item \subsubheaderbf{\cnnavg} A similar approach to \cnnavgI, but with metadata features concatenated to the features of DenseNet before the fully connected layers. 
\item \subsubheaderbf{\lstmI} An LSTM architecture trained using features extracted from \cnnavgI.
\item \subsubheaderbf{\lstm} An LSTM architecture trained using features extracted from \cnnavg.
\end{compactitem}

\begin{table}[!t]
	\centering
	\includegraphics[width=\columnwidth]{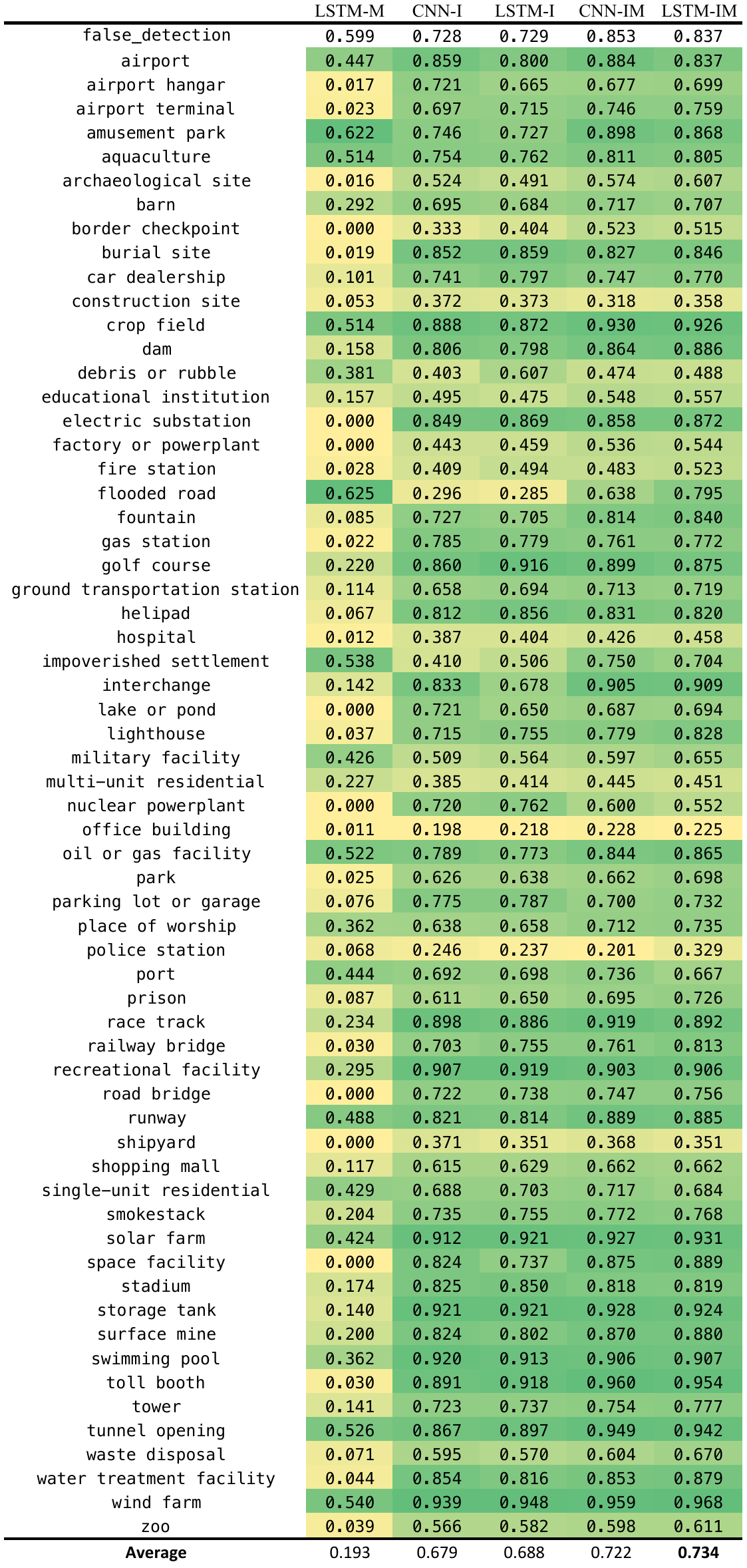}
    \vspace{-0.7cm}
	\caption{F1 scores for different approaches on \test. Color formatting was applied to each column independently. The average values shown at the bottom of the table are calculated without FD scores.}
	\label{tab:baseline_results}
    \vspace{-0.4cm}
\end{table}

\begin{table}[h!]
	\centering
	\setlength{\tabcolsep}{10pt}
	{\small
		\scalebox{0.85}{
			\begin{tabular}{ccccc}
				\toprule  
				\lstmM  & \cnnavgI & \lstmI &  \cnnavg & \lstm \\
				\midrule
				0   & 0.685 & 0.693 & 0.695 & \textbf{0.702} \\
				\bottomrule
			\end{tabular}
		}
	}
    \vspace{-0.2cm}
	\caption{{Results on \test instances where the metadata-only baseline (\lstmM) is not able to correctly predict the category. These are the average F1 scores not including FD. These results show that metadata is important beyond exploiting bias in the dataset.}}
	\label{tab:meta_cannot_predict_table}
\end{table}

\begin{table}[h!]
	\centering
	\setlength{\tabcolsep}{5pt}
	{\small
		\scalebox{0.85}{
			\begin{tabular}{cccccc}
				\toprule  
				\cnnavgI-1 & \cnnavgI & \lstmI & \cnnavg-1 & \cnnavg & \lstm \\
				\midrule
				0.618 & 0.678 & 0.684 & 0.666 & 0.722 & \textbf{0.735} \\
				\bottomrule
			\end{tabular}
		}
	}
	\caption{{Average F1 scores, not including FD, for individual images from \test. \cnnavgI-1 and \cnnavg-1 make predictions for each individual view. All other methods repeat their prediction over the full sequence.}}
	\label{tab:individual_views_results}
\end{table}

The LSTM models, which were also trained with the Adam optimizer~\cite{kingma2014adam}, contained 4096-d hidden states, which were passed to a 512-d multi-layer perceptron (MLP). All of these methods are trained on \train + \val. As tight bounding boxes are typically provided for category instances in the dataset, we add a context buffer around each box before extracting the region of interest from the image. We found that it was useful to provide more context for categories with smaller sizes (\eg, single-unit residential, fountain) and less context for categories that generally cover larger areas (\eg, airports, nuclear power plants).

\begin{figure*}[t!]
	\centering
	\includegraphics[width=\textwidth]{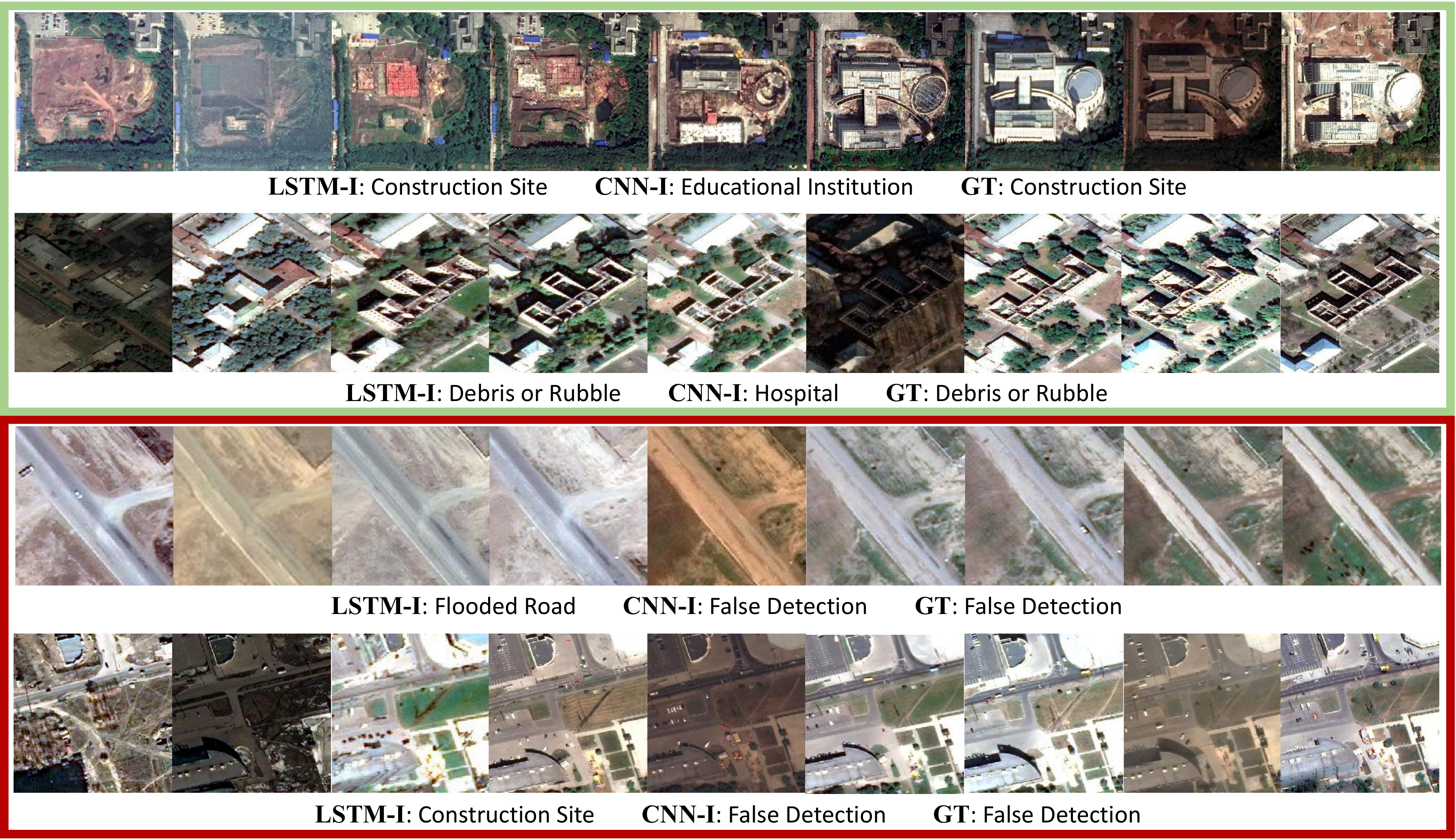}
	\caption{Qualitative examples from \test of the image-only approaches. The images presented here show the extracted and resized images that are passed to the CNN approaches. The top two rows show success cases for \lstmI, where \cnnavgI was not able to correctly predict the category. The bottom two rows show failure cases for \lstmI, where \cnnavgI was able to correctly predict the category. Note that sequences with $\geq$9 views were chosen and additional views were trimmed to keep the figure rectangular.}
	\label{fig:qualitative examples}
\end{figure*}

Per-category F1 scores for \test are shown in Table~\ref{tab:baseline_results}. From the results, it can be observed that, in general, the LSTM architectures show similar performance to our approaches that sum the probabilities over each view. Some possible contributors to this are the large quantity of single-view images provided in the dataset and that temporal changes may not be particularly important for several of the categories. \cnnavgI and \cnnavg are also, to some extent, already reasoning about temporal information while making predictions by summing the softmax outputs over each temporal view. Qualitative results that show success and failure cases for \lstmI are shown in \figref{fig:qualitative examples}. Qualitative results are not shown for the approaches that use metadata, as it is much harder to visually show why the methods succeed in most cases.

It could be argued that the results for approaches using metadata are only making improvements because of bias exploitation. To show that metadata helps beyond inherent bias, we removed all instances from the test set where the metadata-only baseline (\lstmM) is able to correctly predict the category. The results of this removal, which can be found in Table~\ref{tab:meta_cannot_predict_table}, show that metadata can still be useful for improving performance.

To further confirm the importance of temporal reasoning, we compare the methods presented above with two additional methods, \cnnavgI-1 and \cnnavg-1, which make predictions for each individual view. We then have all other methods repeat their prediction over the full sequence. This is done to show that, on average, seeing an area multiple times outperforms single-view predictions. We note that these tests are clearly not fair for some categories, such as ``construction site", where some views may not even contain the category. However, we perform these tests for completeness to confirm our expectations. Results are shown in Table~\ref{tab:individual_views_results}. Per-category results are in the appendix.

%% file: conclusion_discussion.tex
\section{Conclusion and Discussion}
\label{sec:conclusion_discussion}

We present \fmow, a dataset that consists of over 1 million satellite images. Temporal views, multispectral imagery, and metadata are provided to enable new types of joint reasoning. Models may leverage temporal information and simultaneously reason about the rich set of metadata features (\eg, timestamp, UTM zone) provided for each image. By posing a task in between detection and classification, we avoid the inherent challenges associated with collecting a large geographically-diverse detection dataset, while still allowing for models to be trained that are transferable to real-world detection systems. Different methods were presented for this task that demonstrate the importance of joint reasoning about metadata and temporal information. All code, data, and pretrained models have been made publicly available. We hope that by releasing the dataset and code, other researchers in the CV community will find new and interesting ways to further utilize the metadata and temporal changes to a scene. We also hope to see \fmow being used to train models that are able to assist in humanitarian efforts, such as applications involving disaster relief.

\subsubheaderbf{Acknowledgments}
This work would not have been possible without the help of everyone on the \fmow Challenge team, who we thank for their contributions. A special thanks to: Kyle Ellis, Todd Bacastow, Alex Dunmire, and Derick Greyling from DigitalGlobe; Rebecca Allegar, Jillian Brennan, Dan Reitz, and Ian Snyder from Booz Allen Hamilton; Kyle Bowerman and G\H{o}d\'eny Bal\'azs from Topcoder; and, finally, Myron Brown, Philippe Burlina, Alfred Mayalu, and Nicolas Norena Acosta from JHU/APL. We also thank the professors, graduate students, and researchers in industry and from the CV community for their suggestions and participation in discussions that helped shape the direction of this work. 

The material in this paper is based upon work supported by the Office of the Director of National Intelligence (ODNI), Intelligence Advanced Research Projects Activity (IARPA), via Contract 2017-17032700004. The views and conclusions contained herein are those of the authors and should not be interpreted as necessarily representing the official policies or endorsements, either expressed or implied, of the ODNI, IARPA, or the U.S. Government. The U.S. Government is authorized to reproduce and distribute reprints for Governmental purposes notwithstanding any copyright annotation therein.

%% file: appendix.tex
\appendix

\renewcommand{\thesection}{Appendix \Roman{section}} 

\vspace{0.5cm}

\section*{Appendix Overview}

In this document, we provide:

\begin{compactitem}
\item[] \ref{sec:appendix_metadata}: Descriptions  and distributions of metadata features.
\item[] \ref{sec:appendix_collection}: Additional collection details.
\item[] \ref{sec:appendix_results}: Additional results.
\item[] \ref{sec:appendix_examples}: Examples from our dataset.
\end{compactitem}

\vspace{0.5cm}

\section{Metadata Features and Statistics}
\label{sec:appendix_metadata}

\begin{enumerate}
\item \subsubheaderbf{ISO Country Code} ISO Alpha-3 country code (String). There are a total of 247 possible country codes, 207 of which are present in \fmow.
\item \subsubheaderbf{UTM Zone} Universal Transverse Mercator. There are 60 UTM zones, which are 6$^\circ$ in width.
We provide a number for the UTM zone (1-60), along with a letter representing the latitude band. There are a total of 20 latitude bands, which range from ``C'' to ``X'' (``I'' and ``O'' are not included).
\item \subsubheaderbf{Timestamp} UTC timestamp. Datetime format (Python): ``\%Y-\%m-\%dT\%H:\%M:\%SZ'' (String).
\item \subsubheaderbf{Cloud Cover} Fraction of the image strip, not image chip, that is completely obscured by clouds on a scale of 0-100 (Integer).
\item \subsubheaderbf{Scan Direction} The direction the sensor is pointed when collecting an image strip. Either ``Forward", when the image is collected ahead of the orbital path or ``Reverse" when the image is taken behind the orbital path (String).
\item \subsubheaderbf{Pan Resolution} Ground sample distance of panchromatic band (pan-GSD) in the image strip, measured in meters (Double). \texttt{start}, \texttt{end}, \texttt{min}, and \texttt{max} values are also included.  \texttt{start} and \texttt{end} represent the pan-GSD for the first and last scan lines, respectively. \texttt{min} and \texttt{max} represent the minimum and maximum pan-GSD for all scan lines, respectively.
\item \subsubheaderbf{Multi Resolution} Ground sample distance of multispectral bands (multi-GSD) in the image strip, measured in meters (Double). \texttt{start}, \texttt{end}, \texttt{min}, and \texttt{max} values are also included.  \texttt{start} and \texttt{end} represent the multi-GSD for the first and last scan lines, respectively. \texttt{min} and \texttt{max} represent the minimum and maximum multi-GSD for all scan lines, respectively.
\item \subsubheaderbf{Target Azimuth} Azimuth angle of the sensor with respect to the center of the image strip, measured in degrees (Double). \texttt{start}, \texttt{end}, \texttt{min}, and \texttt{max} values are also included.  \texttt{start} and \texttt{end} represent the target azimuth for the first and last scan lines, respectively. \texttt{min} and \texttt{max} represent the minimum and maximum target azimuth for all scan lines, respectively.
\item \subsubheaderbf{Sun Azimuth} Azimuth angle of the sun measured from north, clockwise in degrees, to the center of the image strip, measured in degrees (Double). \texttt{min} and \texttt{max} values are also included. \texttt{min} and \texttt{max} represent the minimum and maximum sun azimuth for all scan lines, respectively.
\item \subsubheaderbf{Sun Elevation} Elevation angle of the sun measured from the horizontal, measured in degrees (Double). \texttt{min} and \texttt{max} values are also included. \texttt{min} and \texttt{max} represent the minimum and maximum sun elevation for all scan lines, respectively.
\item \subsubheaderbf{Off-Nadir Angle} The off nadir angle of the satellite with respect to the center of the image strip, measured in degrees (Double). \texttt{start}, \texttt{end}, \texttt{min}, and \texttt{max} values are also included. \texttt{start} and \texttt{end} represent the off-nadir angle for the first and last scan lines, respectively. \texttt{min} and \texttt{max} represent the minimum and maximum off-nadir angle for all scan lines, respectively.
\end{enumerate}

\vspace{1cm}

\subsubheaderbf{Country Codes}
Here we show the counts for each unique country code in \fmow. Counts are incremented once for each sequence instead of once per metadata file.

[(``USA", 18750), (``FRA", 7470), (``ITA", 6985), (``RUS", 6913), (``CHN", 6597), (``DEU", 4686), (``GBR", 4496), (``BRA", 3820), (``CAN", 3128), (``TUR", 2837), (``JPN", 2542), (``IDN", 2448), (``ESP", 2402), (``AUS", 2105), (``DZA", 1849), (``IND", 1804), (``UKR", 1735), (``CZE", 1713), (``POL", 1386), (``MEX", 1274), (``ARG", 1248), (``NLD", 1236), (``SYR", 1224), (``BEL", 1190), (``PHL", 1179), (``IRQ", 1129), (``EGY", 1041), (``ZAF", 924), (``CHL", 888), (``LTU", 871), (``LBY", 863), (``KOR", 809), (``CHE", 788), (``LVA", 772), (``PRT", 722), (``YEM", 701), (``BLR", 601), (``GRC", 592), (``AUT", 572), (``SVN", 570), (``ARE", 566), (``IRN", 540), (``COL", 509), (``TWN", 509), (``TZA", 475), (``NZL", 465), (``PER", 459), (``HTI", 417), (``KEN", 405), (``NGA", 383), (``VEN", 378), (``PRK", 371), (``ECU", 351), (``IRL", 335), (``MYS", 328), (``BOL", 313), (``FIN", 288), (``KAZ", 268), (``MAR", 266), (``TUN", 257), (``CUB", 256), (``EST", 247), (``SAU", 246), (``HUN", 222), (``THA", 219), (``NPL", 196), (``HRV", 187), (``NOR", 183), (``SVK", 175), (``SEN", 172), (``BGD", 171), (``HND", 167), (``SWE", 166), (``BGR", 165), (``HKG", 154), (``DNK", 153), (``MDA", 147), (``ROU", 142), (``ZWE", 141), (``SRB", 140), (``GTM", 140), (``DOM", 134), (``LUX", 133), (``SDN", 132), (``VNM", 126), (``URY", 120), (``CRI", 119), (``SOM", 112), (``ISL", 110), (``LKA", 110), (``QAT", 108), (``PRY", 107), (``SGP", 106), (``OMN", 105), (``PRI", 95), (``NIC", 87), (``NER", 85), (``SSD", 82), (``UGA", 79), (``SLV", 79), (``JOR", 78), (``CMR", 77), (``PAN", 74), (``PAK", 72), (``UZB", 70), (``CYP", 67), (``KWT", 67), (``ALB", 66), (``CIV", 65), (``BHR", 65), (``GIN", 64), (``MLT", 63), (``JAM", 62), (``AZE", 62), (``GEO", 60), (``SLE", 59), (``ETH", 58), (``LBN", 57), (``ZMB", 55), (``TTO", 54), (``LBR", 52), (``BWA", 51), (``ANT", 50), (``BHS", 50), (``MNG", 46), (``MKD", 45), (``GLP", 45), (``COD", 45), (``KO-", 42), (``BEN", 42), (``GHA", 41), (``MDG", 36), (``MLI", 35), (``AFG", 35), (``ARM", 33), (``MRT", 33), (``KHM", 32), (``CPV", 31), (``TKM", 31), (``MMR", 31), (``BFA", 29), (``BLZ", 29), (``NCL", 28), (``AGO", 27), (``FJI", 26), (``TCD", 25), (``MTQ", 25), (``GMB", 23), (``SWZ", 23), (``BIH", 21), (``CAF", 19), (``GUF", 19), (``PSE", 19), (``MOZ", 18), (``NAM", 18), (``SUR", 17), (``GAB", 17), (``LSO", 16), (``ERI", 15), (``BRN", 14), (``REU", 14), (``GUY", 14), (``MAC", 13), (``TON", 13), (``ABW", 12), (``PYF", 12), (``TGO", 12), (``BRB", 12), (``VIR", 11), (``CA-", 11), (``DJI", 11), (``FLK", 11), (``MNE", 11), (``KGZ", 11), (``ESH", 10), (``LCA", 10), (``BMU", 10), (``COG", 9), (``ATG", 9), (``BDI", 9), (``GIB", 8), (``LAO", 8), (``GNB", 8), (``DMA", 8), (``KNA", 8), (``GNQ", 7), (``RWA", 7), (``BTN", 7), (``TJK", 6), (``TCA", 5), (``VCT", 4), (``WSM", 3), (``IOT", 3), (``AND", 3), (``ISR", 3), (``AIA", 3), (``MDV", 2), (``SHN", 2), (``VGB", 2), (``MSR", 2), (``PNG", 1), (``MHL", 1), (``VUT", 1), (``GRD", 1), (``VAT", 1), (``MCO", 1)]

\vspace{1cm}

\subsubheaderbf{UTM Zones}
Here we show the counts for each unique UTM zone in \fmow. Counts are incremented once for each sequence instead of once per metadata file.

[(``31U", 5802), (``32T", 4524), (``33T", 4403), (``30U", 4186), (``32U", 3864), (``33U", 3315), (``31T", 3150), (``18T", 2672), (``17T", 2339), (``34U", 2049), (``37S", 1718), (``30T", 1686), (``37U", 1672), (``23K", 1627), (``18S", 1481), (``11S", 1388), (``16T", 1283), (``54S", 1244), (``38S", 1229), (``31S", 1227), (``35U", 1137), (``35V", 1116), (``52S", 1115), (``16S", 1110), (``51P", 1086), (``51R", 1069), (``36S", 1046), (``35T", 1038), (``36R", 1037), (``49M", 1026), (``48M", 1021), (``10T", 1010), (``53S", 1001), (``10S", 955), (``14R", 935), (``19T", 928), (``30S", 912), (``17S", 875), (``17R", 874), (``43P", 854), (``50S", 796), (``36U", 767), (``50R", 751), (``33S", 751), (``32S", 746), (``14S", 730), (``34T", 728), (``12S", 716), (``37M", 705), (``13S", 676), (``37T", 667), (``36T", 653), (``15S", 629), (``55H", 618), (``34S", 604), (``29S", 600), (``38P", 598), (``15T", 586), (``22J", 585), (``18Q", 549), (``15R", 539), (``35S", 511), (``10U", 497), (``21H", 492), (``36V", 491), (``19H", 482), (``48R", 476), (``49S", 459), (``48S", 446), (``49Q", 444), (``29T", 438), (``16P", 429), (``56H", 425), (``14Q", 422), (``40R", 420), (``39R", 413), (``39U", 406), (``18N", 385), (``35J", 383), (``37V", 380), (``50T", 379), (``56J", 355), (``34V", 351), (``43V", 347), (``29U", 346), (``38U", 345), (``17M", 328), (``38T", 323), (``19P", 323), (``51S", 317), (``54H", 311), (``49R", 295), (``34H", 293), (``22K", 293), (``48N", 276), (``20H", 273), (``50Q", 268), (``28P", 262), (``18L", 260), (``24M", 258), (``24L", 256), (``21J", 255), (``41V", 254), (``13T", 254), (``47N", 253), (``40U", 253), (``45R", 251), (``43Q", 245), (``51Q", 243), (``51T", 240), (``39S", 239), (``19K", 238), (``19Q", 237), (``59G", 236), (``43R", 234), (``12T", 230), (``49T", 227), (``41U", 223), (``32V", 219), (``30V", 212), (``13Q", 212), (``40V", 210), (``16R", 210), (``20T", 210), (``38R", 204), (``36J", 203), (``46T", 200), (``45T", 197), (``44U", 196), (``15Q", 190), (``50L", 190), (``32P", 184), (``60H", 182), (``47P", 182), (``20P", 181), (``24K", 178), (``17Q", 178), (``35K", 169), (``20J", 168), (``11U", 165), (``18H", 164), (``52T", 163), (``11T", 161), (``36N", 158), (``39V", 157), (``20K", 157), (``39Q", 155), (``12U", 149), (``38V", 147), (``18P", 147), (``23L", 147), (``18G", 146), (``31N", 146), (``19J", 142), (``33P", 141), (``40Q", 136), (``13R", 136), (``47T", 132), (``47R", 126), (``48U", 124), (``32R", 123), (``15P", 121), (``39P", 117), (``48P", 117), (``33R", 116), (``45U", 113), (``43S", 111), (``44N", 109), (``54T", 109), (``32N", 109), (``36W", 108), (``17P", 108), (``36P", 105), (``31R", 104), (``56K", 101), (``20Q", 101), (``39T", 97), (``16Q", 96), (``29R", 95), (``25L", 92), (``45Q", 91), (``46Q", 91), (``48T", 90), (``44Q", 89), (``42V", 87), (``29N", 87), (``43U", 86), (``4Q", 86), (``47Q", 85), (``48Q", 84), (``30N", 83), (``19G", 82), (``25M", 81), (``42Q", 80), (``44P", 80), (``20L", 77), (``50J", 77), (``53U", 76), (``38N", 75), (``27W", 75), (``44R", 75), (``33V", 74), (``34R", 72), (``49L", 70), (``36M", 69), (``40S", 69), (``12R", 68), (``37P", 68), (``52R", 65), (``14T", 64), (``50U", 62), (``35H", 62), (``50H", 61), (``28R", 60), (``54U", 59), (``46V", 58), (``44T", 56), (``21K", 56), (``55G", 56), (``22L", 56), (``35P", 55), (``31P", 54), (``29P", 54), (``35R", 52), (``30R", 51), (``19U", 50), (``53T", 49), (``46U", 49), (``50N", 48), (``47S", 48), (``42R", 48), (``37Q", 47), (``19L", 47), (``14U", 47), (``28Q", 46), (``37N", 45), (``19F", 45), (``42U", 44), (``36K", 42), (``37R", 40), (``37W", 40), (``41S", 38), (``42S", 38), (``38Q", 37), (``30P", 37), (``42T", 36), (``35L", 36), (``46R", 36), (``52U", 35), (``60G", 35), (``27V", 34), (``45V", 34), (``35W", 34), (``13U", 34), (``35M", 34), (``18M", 32), (``17L", 32), (``41W", 32), (``17N", 31), (``21N", 31), (``23M", 30), (``21L", 29), (``28S", 28), (``58K", 28), (``22M", 28), (``41R", 27), (``18R", 27), (``10V", 26), (``57U", 26), (``34K", 26), (``49U", 25), (``6V", 25), (``38L", 25), (``20G", 25), (``33L", 24), (``60K", 24), (``55K", 23), (``51N", 23), (``22H", 22), (``22N", 22), (``47V", 22), (``41T", 21), (``44V", 21), (``36Q", 21), (``46S", 20), (``22T", 20), (``34N", 19), (``20U", 19), (``12Q", 19), (``12V", 19), (``19N", 18), (``31Q", 18), (``21M", 18), (``52L", 18), (``56V", 18), (``52V", 18), (``23J", 16), (``45W", 16), (``9U", 16), (``34J", 16), (``27P", 16), (``43W", 15), (``1K", 14), (``33M", 14), (``40W", 14), (``40K", 14), (``43T", 14), (``55T", 14), (``51U", 13), (``53K", 13), (``34M", 13), (``32M", 13), (``37L", 13), (``21P", 12), (``50P", 12), (``35N", 12), (``6K", 11), (``59H", 11), (``33K", 11), (``20M", 11), (``49N", 11), (``5Q", 10), (``6W", 10), (``26Q", 10), (``39L", 10), (``47U", 10), (``34W", 10), (``50K", 10), (``8V", 10), (``20S", 10), (``40T", 9), (``51V", 9), (``42W", 8), (``60W", 8), (``53H", 8), (``50V", 8), (``20F", 8), (``53L", 7), (``18F", 7), (``35Q", 7), (``30Q", 7), (``44S", 7), (``15M", 7), (``5V", 7), (``54J", 7), (``39W", 6), (``49P", 6), (``50M", 6), (``19V", 6), (``21F", 6), (``20N", 5), (``14P", 5), (``34P", 5), (``53J", 5), (``38M", 5), (``51K", 5), (``29Q", 4), (``11R", 4), (``49V", 4), (``48V", 4), (``51M", 4), (``38W", 4), (``33N", 4), (``45S", 4), (``27Q", 4), (``55J", 3), (``19M", 3), (``53V", 3), (``2W", 3), (``32Q", 3), (``2L", 3), (``16M", 3), (``57W", 3), (``43M", 3), (``53W", 2), (``43N", 2), (``52J", 2), (``28M", 2), (``56T", 2), (``33H", 2), (``21T", 2), (``44W", 2), (``15V", 1), (``33W", 1), (``60V", 1), (``18K", 1), (``31M", 1), (``54M", 1), (``58P", 1), (``58W", 1), (``40X", 1), (``58G", 1), (``57V", 1), (``16U", 1), (``59K", 1), (``52N", 1), (``2K", 1), (``33Q", 1), (``34Q", 1), (``11V", 1), (``56W", 1), (``26P", 1), (``28W", 1), (``59W", 1), (``38K", 1), (``26S", 1), (``7L", 1), (``56U", 1), (``55V", 1)]

\vspace{0.5cm}

\section{Dataset Collection}
\label{sec:appendix_collection}

\begin{figure}[h!]
    \centering
    \includegraphics[width=\columnwidth]{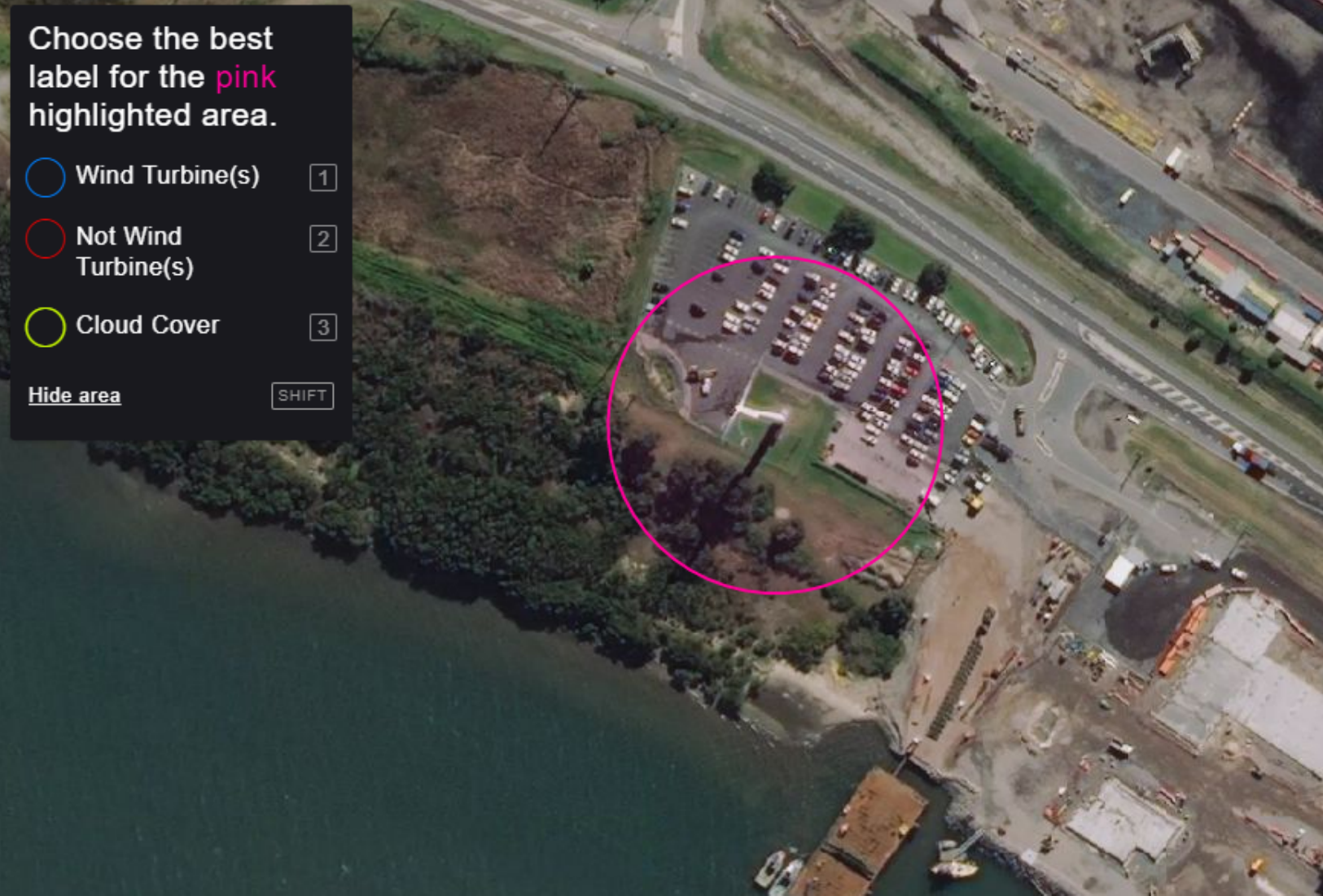}
	\caption{Sample image (``wind farm") of what a GeoHIVE user might see while validating potential \fmow features. Instructions can be seen in the top-left corner that inform users to press the `1', `2', or `3' keys to validate existence, non-existence, or cloud obscuration of a particular object.} 
    \label{fig:GeoHive_existence}
\end{figure}

The location selection phase was used to identify potential locations that map to our categories while also ensuring geographic diversity. Potential locations were drawn from several Volunteered Geographic Information (VGI) datasets, which were conflated and curated to remove duplicates and ensure geographic diversity. The remaining locations were then processed using DigitalGlobe's GeoHIVE crowdsourcing platform. Members of the GeoHIVE crowd were asked to validate the presence of categories in satellite images, as shown in \figref{fig:GeoHive_existence}. The interface uses center-point location information to draw a circle around a possible object of interest. The interface then asks users to rapidly verify the existence of a particular label, as extracted from the VGI datasets, using the `1', `2', and `3' keys to represent existence, non-existence, and cloud cover.

\vspace{1.5cm}

\vspace{1.5cm}

\begin{figure}[h!]
	\begin{subfigure}{\columnwidth}
		\centering
		\includegraphics[width=\columnwidth]{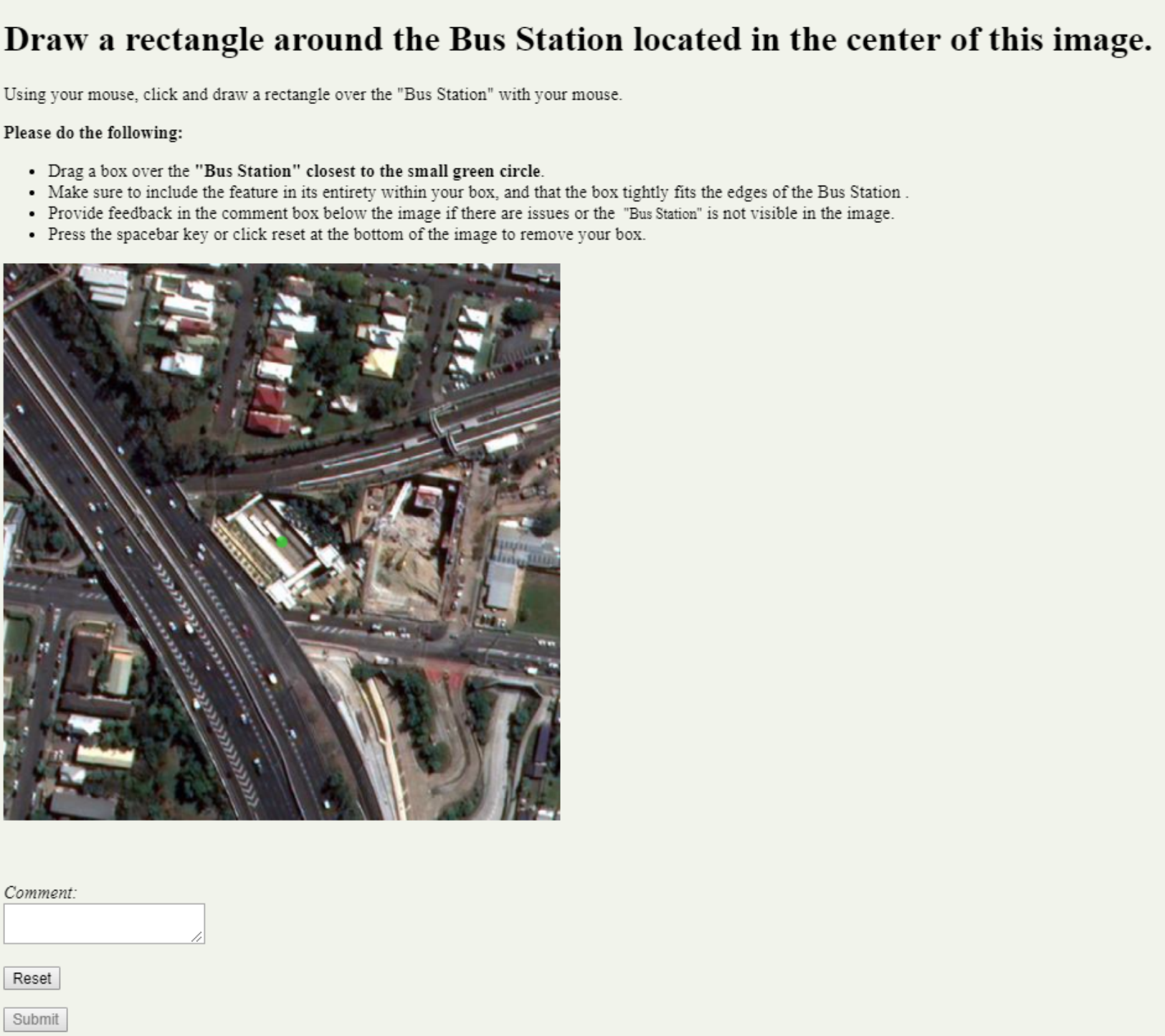}
		\caption{ground transportation station} \label{fig:GeoHive_localization_busstation}
	\end{subfigure}

	\vspace{1.5cm}

	\begin{subfigure}{\columnwidth}
		\centering
		\includegraphics[width=\columnwidth]{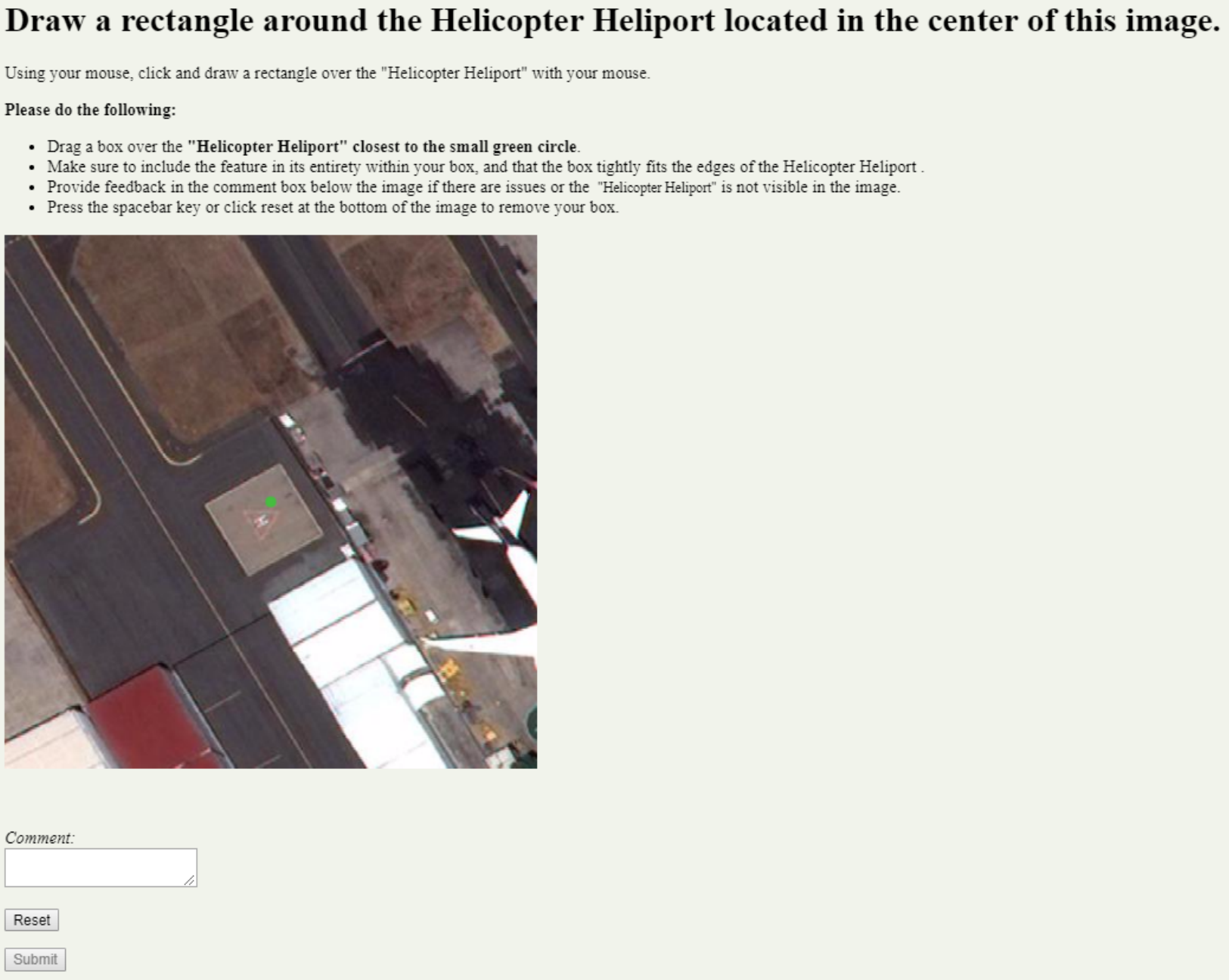}
		\caption{helipad} \label{fig:GeoHive_localization_helipad}
	\end{subfigure}
	\caption{Sample images of the interface used to more precisely localize objects within an image. In each example, a green dot is placed near the center of the pertinent object. Users are able to draw a bounding box by clicking and dragging. Instructions at the top of each example inform the user how to use the interface and also provide any category-specific instructions that may be relevant. Comments regarding issues such as clouds or object misclassification can be entered near the bottom of the page before submitting an annotation.} \label{fig:GeoHive_localization}
\end{figure}

For validation of object localization, a different interface is used that asks users to draw a bounding box around the object of interest after being given an initial seed point. The visualization for this is shown in \figref{fig:GeoHive_localization}, and the seed point can be seen as a green dot located on the object of interest. Users are additionally provided some instructions regarding how large of a box to draw, which may vary by object class. This interface is more complex than the location selection interface, which is why it is performed after object existence can be confirmed and non-cloudy high-quality imagery is obtained. A smaller and more experienced group of users is also used for this task to help ensure the quality of the annotations.

\begin{figure}[h!]
    \centering
    \includegraphics[width=\columnwidth]{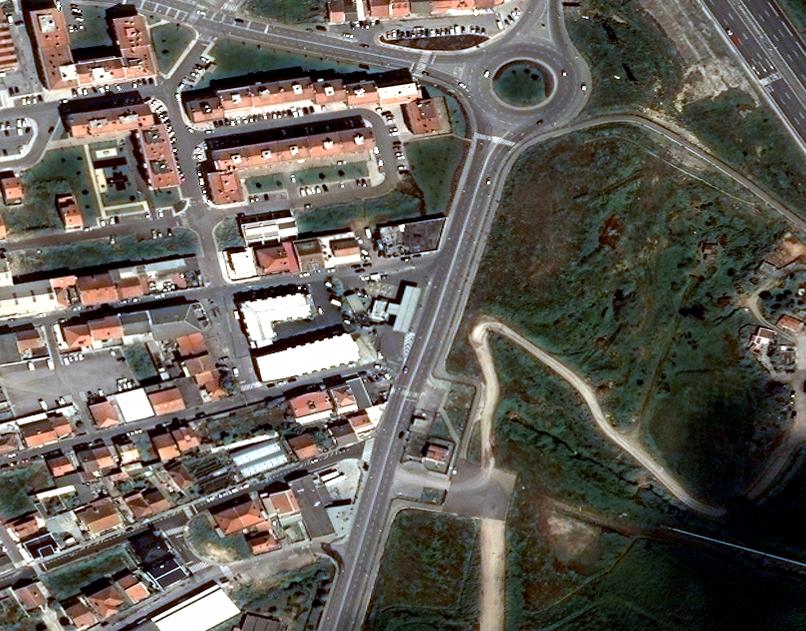}
	\caption{This image shows an example from \fmow with an instance of the ``gas station'' category, which is located near the center of the image. This shows how it is difficult to identify the functional purpose of the surrounding buildings. If map data is not available, it is very easy for humans to make mistakes when annotating. } 
    \label{fig:gas_station_annotation_difficulty}
\end{figure}

To help illustrate why full image annotation of \fmow categories is difficult, we show an example from the dataset in \figref{fig:gas_station_annotation_difficulty}. The primary category, which is located near the center of the image, is ``gas station''. As shown, it is difficult to identify the functional purpose of the surrounding buildings, and if map data is not available, it would be easy for humans to make mistakes when annotating. It is also possible to see how object detectors may correctly detect other categories from \fmow. By providing bounding boxes as input, we can avoid the issue of scoring results for which annotations do not exist.

\vspace{1.5cm}

\vspace{1cm}

\begin{figure}[h!]
    \centering
    \includegraphics[width=\columnwidth]{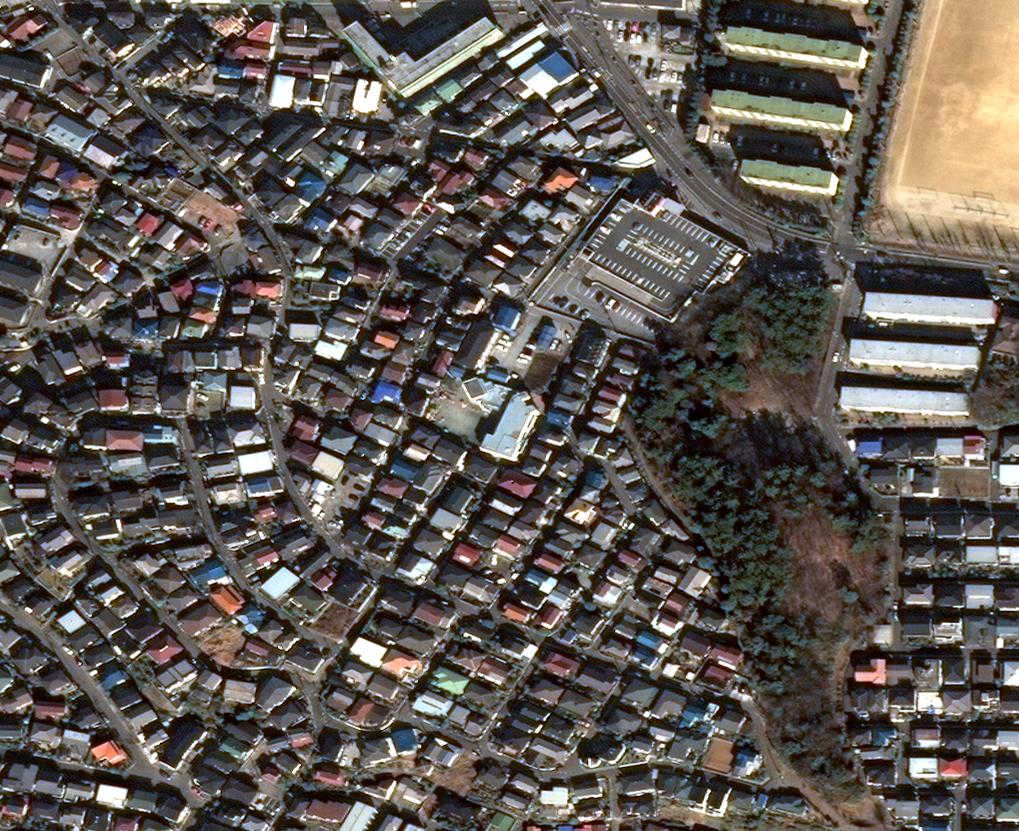}
	\caption{This image shows an example from \fmow with an instance of the ``educational institution'' category. This example is located in Japan. This shows the difficulty of determining which buildings/areas should be included within the bounding box as part of the category. } 
    \label{fig:educational_institution_annotation_difficulty}
\end{figure}

Another difficult example is shown in \figref{fig:educational_institution_annotation_difficulty}, which shows an instance of the ``educational institution'' category in Japan. While an initial box is provided to the annotators, it is difficult to determine which buildings should be grouped as part of the same category when placing a bounding box. 
\vspace{0.5cm}

\section{Additional Results}
\label{sec:appendix_results}

Introduced in the main paper, \cnnavgI-1 and \cnnavg-1 make predictions for each individual view. All other methods repeat their prediction over the full sequence. Again, we note that these tests are clearly not fair to some categories, such as ``construction site'', where some views may not even contain the category. However, we show results of these tests for completeness. Only the average values, which do not include ``false detection'' results, are shown in the main paper. We show per-category results in Table~\ref{tab:individual_baseline_results_table}.

\begin{table}[h!]
    \centering
    \includegraphics[width=\columnwidth]{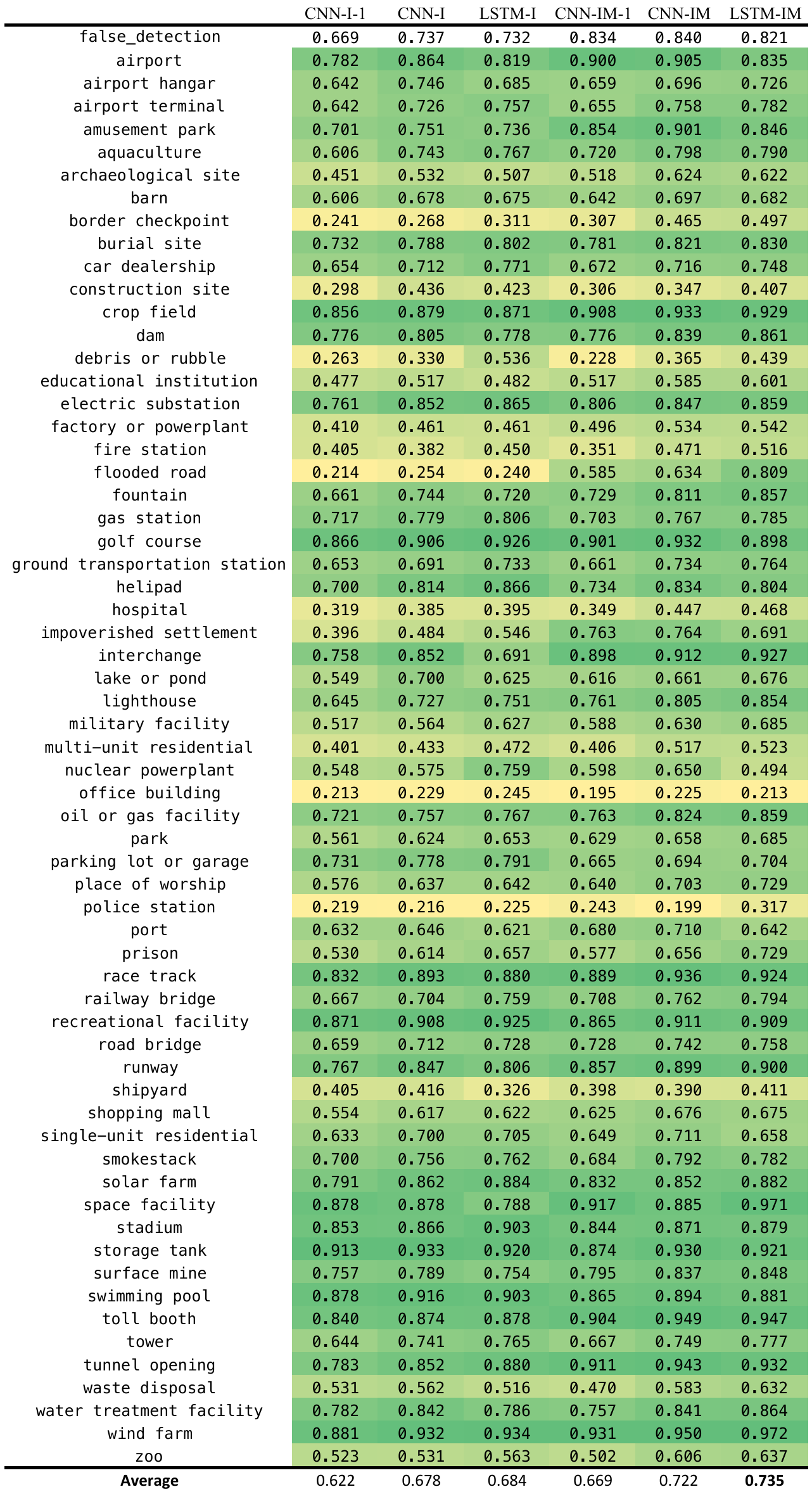}
    \caption{F1 scores for different approaches on an individual image basis. Color formatting was applied to each column independently. The average values shown at the bottom of the table are calculated without the false detection scores. \cnnavgI-1 and \cnnavg-1 make predictions for each individual view. All other methods repeat their prediction over the full sequence.}
    \label{tab:individual_baseline_results_table}
\end{table}

\section{Dataset Examples}
\label{sec:appendix_examples}

\begin{figure*}[h!]

\centering

\stackunder[1pt]{\includegraphics[width=0.11\textwidth]{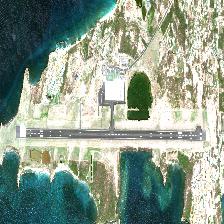}}{\scriptsize airport}
\hspace{0.01cm}
\stackunder[1pt]{\includegraphics[width=0.11\textwidth]{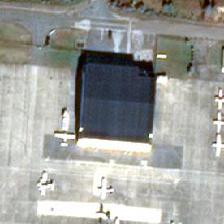}}{\scriptsize airport hangar}
\hspace{0.01cm}
\stackunder[1pt]{\includegraphics[width=0.11\textwidth]{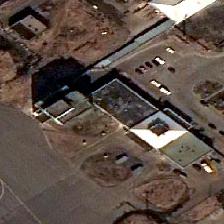}}{\scriptsize airport terminal}
\hspace{0.01cm}
\stackunder[1pt]{\includegraphics[width=0.11\textwidth]{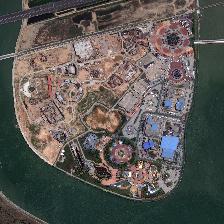}}{\scriptsize amusement park}
\hspace{0.01cm}
\stackunder[1pt]{\includegraphics[width=0.11\textwidth]{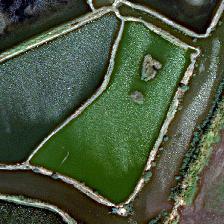}}{\scriptsize aquaculture}
\hspace{0.01cm}
\stackunder[1pt]{\includegraphics[width=0.11\textwidth]{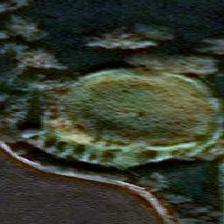}}{\scriptsize archaeological site}
\hspace{0.01cm}
\stackunder[1pt]{\includegraphics[width=0.11\textwidth]{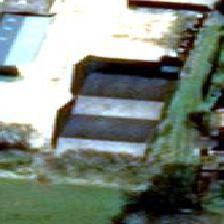}}{\scriptsize barn}
\hspace{0.01cm}
\stackunder[1pt]{\includegraphics[width=0.11\textwidth]{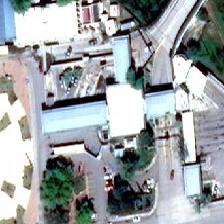}}{\scriptsize border checkpoint}

\stackunder[1pt]{\includegraphics[width=0.11\textwidth]{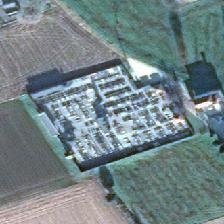}}{\scriptsize burial site}
\hspace{0.01cm}
\stackunder[1pt]{\includegraphics[width=0.11\textwidth]{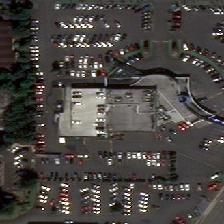}}{\scriptsize car dealership}
\hspace{0.01cm}
\stackunder[1pt]{\includegraphics[width=0.11\textwidth]{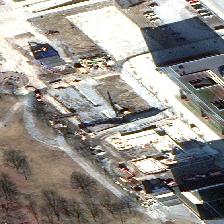}}{\scriptsize construction site}
\hspace{0.01cm}
\stackunder[1pt]{\includegraphics[width=0.11\textwidth]{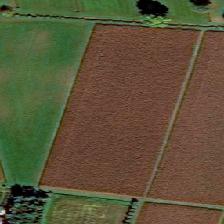}}{\scriptsize crop field}
\hspace{0.01cm}
\stackunder[1pt]{\includegraphics[width=0.11\textwidth]{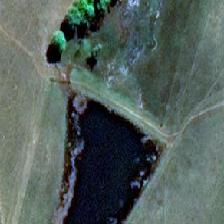}}{\scriptsize dam}
\hspace{0.01cm}
\stackunder[1pt]{\includegraphics[width=0.11\textwidth]{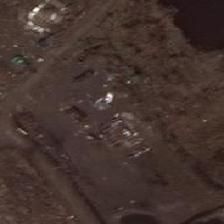}}{\scriptsize debris or rubble}
\hspace{0.01cm}
\stackunder[1pt]{\includegraphics[width=0.11\textwidth]{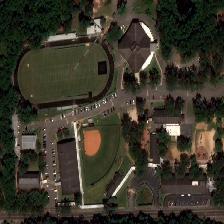}}{\scriptsize educational institution}
\hspace{0.01cm}
\stackunder[1pt]{\includegraphics[width=0.11\textwidth]{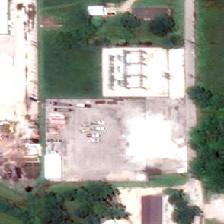}}{\scriptsize electric substation}

\stackunder[1pt]{\includegraphics[width=0.11\textwidth]{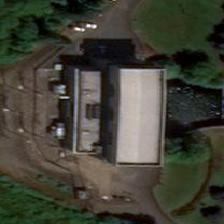}}{\scriptsize factory or powerplant}
\hspace{0.01cm}
\stackunder[1pt]{\includegraphics[width=0.11\textwidth]{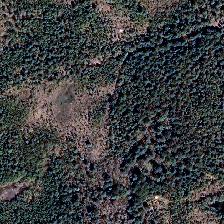}}{\scriptsize false detection}
\hspace{0.01cm}
\stackunder[1pt]{\includegraphics[width=0.11\textwidth]{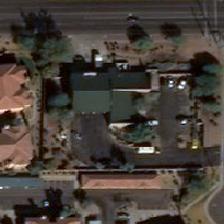}}{\scriptsize fire station}
\hspace{0.01cm}
\stackunder[1pt]{\includegraphics[width=0.11\textwidth]{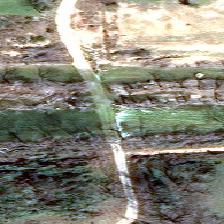}}{\scriptsize flooded road}
\hspace{0.01cm}
\stackunder[1pt]{\includegraphics[width=0.11\textwidth]{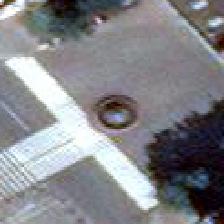}}{\scriptsize fountain}
\hspace{0.01cm}
\stackunder[1pt]{\includegraphics[width=0.11\textwidth]{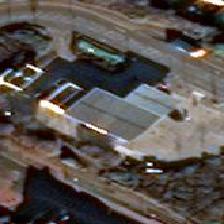}}{\scriptsize gas station}
\hspace{0.01cm}
\stackunder[1pt]{\includegraphics[width=0.11\textwidth]{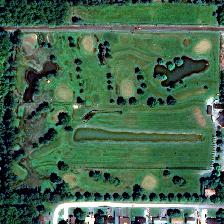}}{\scriptsize golf course}
\hspace{0.01cm}
\stackunder[1pt]{\includegraphics[width=0.11\textwidth]{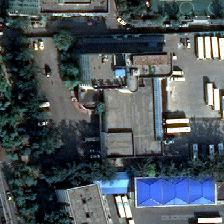}}{\scriptsize    \begin{tabular}{cc}
    ground transportation\\
    station
   \end{tabular} }

\stackunder[1pt]{\includegraphics[width=0.11\textwidth]{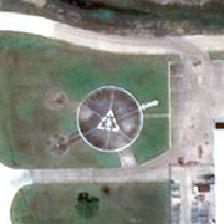}}{\scriptsize helipad}
\hspace{0.01cm}
\stackunder[1pt]{\includegraphics[width=0.11\textwidth]{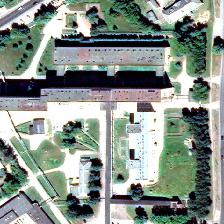}}{\scriptsize hospital}
\hspace{0.01cm}
\stackunder[1pt]{\includegraphics[width=0.11\textwidth]{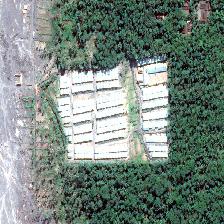}}{\scriptsize impoverished settlement}
\hspace{0.01cm}
\stackunder[1pt]{\includegraphics[width=0.11\textwidth]{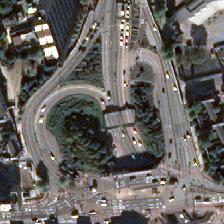}}{\scriptsize interchange}
\hspace{0.01cm}
\stackunder[1pt]{\includegraphics[width=0.11\textwidth]{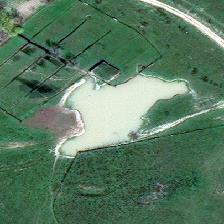}}{\scriptsize lake or pond}
\hspace{0.01cm}
\stackunder[1pt]{\includegraphics[width=0.11\textwidth]{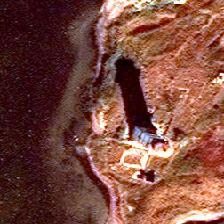}}{\scriptsize lighthouse}
\hspace{0.01cm}
\stackunder[1pt]{\includegraphics[width=0.11\textwidth]{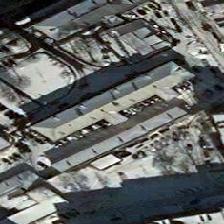}}{\scriptsize military facility}
\hspace{0.01cm}
\stackunder[1pt]{\includegraphics[width=0.11\textwidth]{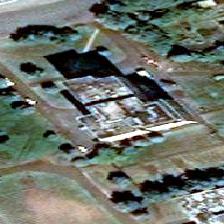}}{\scriptsize multi-unit residential}

\stackunder[1pt]{\includegraphics[width=0.11\textwidth]{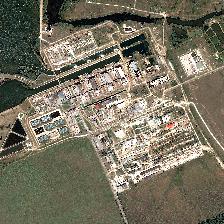}}{\scriptsize nuclear powerplant}
\hspace{0.01cm}
\stackunder[1pt]{\includegraphics[width=0.11\textwidth]{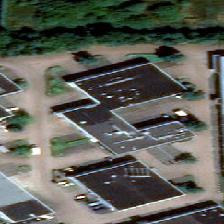}}{\scriptsize office building}
\hspace{0.01cm}
\stackunder[1pt]{\includegraphics[width=0.11\textwidth]{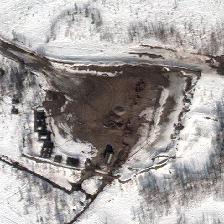}}{\scriptsize oil or gas facility}
\hspace{0.01cm}
\stackunder[1pt]{\includegraphics[width=0.11\textwidth]{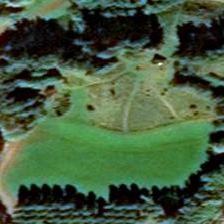}}{\scriptsize park}
\hspace{0.01cm}
\stackunder[1pt]{\includegraphics[width=0.11\textwidth]{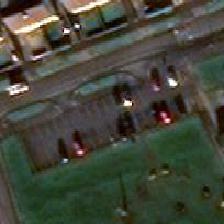}}{\scriptsize parking lot or garage}
\hspace{0.01cm}
\stackunder[1pt]{\includegraphics[width=0.11\textwidth]{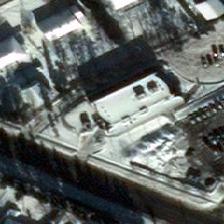}}{\scriptsize place of worship}
\hspace{0.01cm}
\stackunder[1pt]{\includegraphics[width=0.11\textwidth]{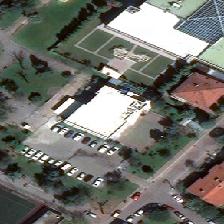}}{\scriptsize police station}
\hspace{0.01cm}
\stackunder[1pt]{\includegraphics[width=0.11\textwidth]{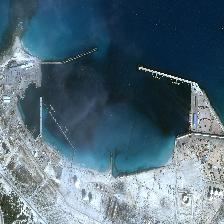}}{\scriptsize port}

\stackunder[1pt]{\includegraphics[width=0.11\textwidth]{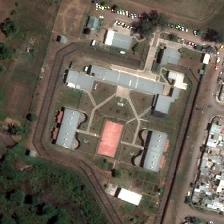}}{\scriptsize prison}
\hspace{0.01cm}
\stackunder[1pt]{\includegraphics[width=0.11\textwidth]{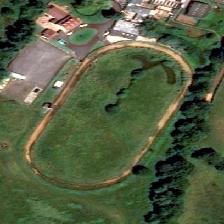}}{\scriptsize race track}
\hspace{0.01cm}
\stackunder[1pt]{\includegraphics[width=0.11\textwidth]{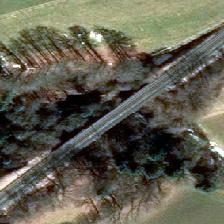}}{\scriptsize railway bridge}
\hspace{0.01cm}
\stackunder[1pt]{\includegraphics[width=0.11\textwidth]{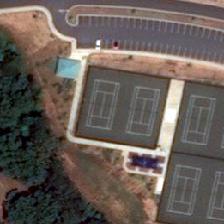}}{\scriptsize recreational facility}
\hspace{0.01cm}
\stackunder[1pt]{\includegraphics[width=0.11\textwidth]{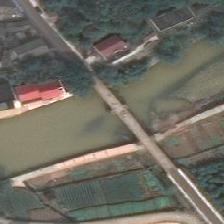}}{\scriptsize road bridge}
\hspace{0.01cm}
\stackunder[1pt]{\includegraphics[width=0.11\textwidth]{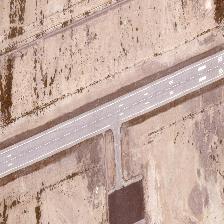}}{\scriptsize runway}
\hspace{0.01cm}
\stackunder[1pt]{\includegraphics[width=0.11\textwidth]{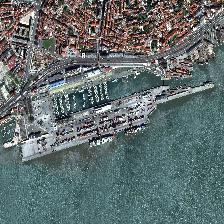}}{\scriptsize shipyard}
\hspace{0.01cm}
\stackunder[1pt]{\includegraphics[width=0.11\textwidth]{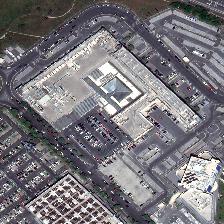}}{\scriptsize shopping mall}

\stackunder[1pt]{\includegraphics[width=0.11\textwidth]{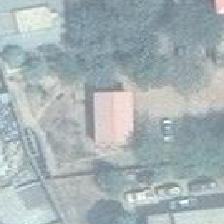}}{\scriptsize single-unit residential}
\hspace{0.01cm}
\stackunder[1pt]{\includegraphics[width=0.11\textwidth]{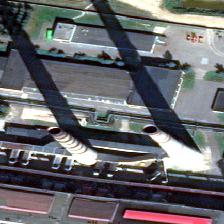}}{\scriptsize smokestack}
\hspace{0.01cm}
\stackunder[1pt]{\includegraphics[width=0.11\textwidth]{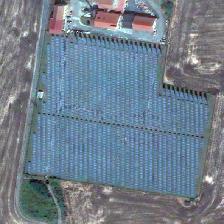}}{\scriptsize solar farm}
\hspace{0.01cm}
\stackunder[1pt]{\includegraphics[width=0.11\textwidth]{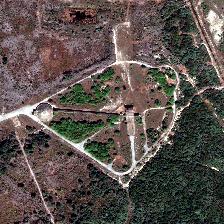}}{\scriptsize space facility}
\hspace{0.01cm}
\stackunder[1pt]{\includegraphics[width=0.11\textwidth]{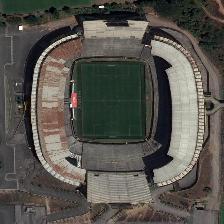}}{\scriptsize stadium}
\hspace{0.01cm}
\stackunder[1pt]{\includegraphics[width=0.11\textwidth]{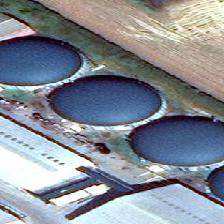}}{\scriptsize storage tank}
\hspace{0.01cm}
\stackunder[1pt]{\includegraphics[width=0.11\textwidth]{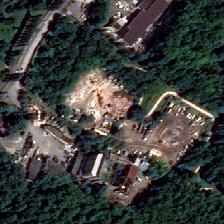}}{\scriptsize surface mine}
\hspace{0.01cm}
\stackunder[1pt]{\includegraphics[width=0.11\textwidth]{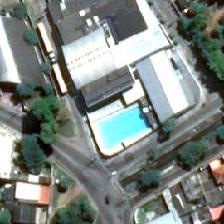}}{\scriptsize swimming pool}

\stackunder[1pt]{\includegraphics[width=0.11\textwidth]{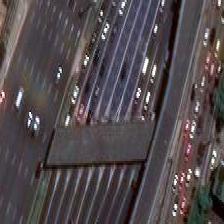}}{\scriptsize toll booth}
\hspace{0.01cm}
\stackunder[1pt]{\includegraphics[width=0.11\textwidth]{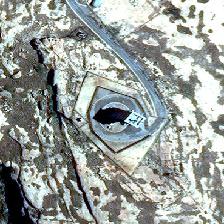}}{\scriptsize tower}
\hspace{0.01cm}
\stackunder[1pt]{\includegraphics[width=0.11\textwidth]{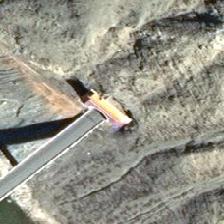}}{\scriptsize tunnel opening}
\hspace{0.01cm}
\stackunder[1pt]{\includegraphics[width=0.11\textwidth]{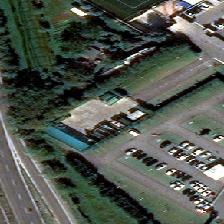}}{\scriptsize waste disposal}
\hspace{0.01cm}
\stackunder[1pt]{\includegraphics[width=0.11\textwidth]{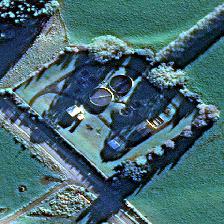}}{\scriptsize water treatment facility}
\hspace{0.01cm}
\stackunder[1pt]{\includegraphics[width=0.11\textwidth]{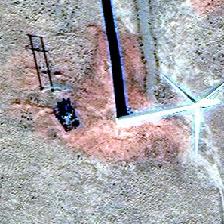}}{\scriptsize wind farm}
\hspace{0.01cm}
\stackunder[1pt]{\includegraphics[width=0.11\textwidth]{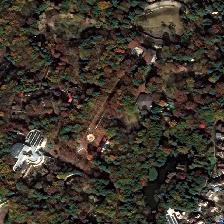}}{\scriptsize zoo}
\hspace{0.01cm}

\caption{One example per category in \fmow.}
\label{fig:dataset_examples}
\end{figure*}

\figref{fig:dataset_examples} shows one example for each category in our dataset. For viewing purposes, regions within the full image chip were extracted using the scaled bounding box coordinates for the categories. For the baseline approaches presented in the main paper, smaller boxes were given more context than larger boxes. Therefore, for smaller-sized categories (\eg, smoke stacks) it may appear that there is a lot more context than expected. It is important to keep in mind that the images for each category in the full dataset vary in quality, recognizability, were taken under different weather conditions (\eg, snow cover) and seasons, contain drastically different context (\eg, desert vs. urban), and other variations. 

\clearpage

\clearpage